\relax
\documentclass[11pt, 3p]{elsarticle}
\usepackage{times}
\usepackage{helvet}
\usepackage{courier}
\usepackage{url}
\usepackage{graphicx}
\usepackage{algorithm}
\usepackage{algorithmicx}
\usepackage[noend]{algpseudocode}
\usepackage{amsmath}
\usepackage{amssymb}
\usepackage{xpatch}
\usepackage{soul}
\usepackage{graphicx}
\usepackage{import}
\usepackage{url}
\usepackage{hyperref}
\usepackage[usenames,dvipsnames]{color}
\usepackage{pgfplots}
\usepackage[per-mode=symbol]{siunitx}
\usepackage[caption=true,labelformat=parens]{subfig}
\usepackage{tikz}

\usetikzlibrary{fit,calc}
\newcommand*{\tikzmk}[1]{\tikz[remember picture,overlay,] \node (#1) {};\ignorespaces}
\newcommand{\boxit}[1]{\tikz[remember picture,overlay]{\node[fill=#1,opacity=.25,xshift=6pt,fit={($(A)-(0.35,0.5\baselineskip)$)($(B)-(0.375\textwidth,0.0)$)}] {};}\ignorespaces}

\newcommand{\boxitone}[1]{\tikz[remember picture,overlay]{\node[fill=#1,opacity=.25,xshift=-3pt,yshift=2pt,fit={($(A)-(0.44,0.5\baselineskip)$)($(B)+(0.945\textwidth,0.0)$)}] {};}\ignorespaces}

\newcommand{\boxitthree}[1]{\tikz[remember picture,overlay]{\node[fill=#1,opacity=.25,xshift=-3pt,yshift=122pt,fit={($(A)-(0.415,0.5\baselineskip)$)($(B)+(0.945\textwidth,0.95)$)}] {};}\ignorespaces}

\newcommand{\boxitfour}[1]{\tikz[remember picture,overlay]{\node[fill=#1,opacity=.25,xshift=286pt,fit={($(A)-(0.0,0.5\baselineskip)$)($(B)-(0.763\textwidth,0.0)$)}] {};}\ignorespaces}

\colorlet{pink}{red!40}
\colorlet{cyan}{cyan!60}

\usepackage{color}
\usepackage{xspace}
\usepackage{enumitem}
\usepackage{xcolor}
\usepackage{sidecap}

\definecolor{codehighlight}{rgb}{0.95,0.8,0.8}
\definecolor{codebackground}{rgb}{0.95,0.95,0.95}

\definecolor{complexbox}{rgb}{0.42,0.83,0.84}
\definecolor{trivialbox}{rgb}{0.98,0.64,0.61}
\definecolor{complexscatter}{rgb}{0.54,0.059,0}
\definecolor{trivialscatter}{rgb}{0.0039,0.09,0.55}
\definecolor{verycomplexscatter}{rgb}{0.0,0.0,0.0}

\definecolor{orange}{rgb}{0.8,0.4,0}
\definecolor{mylink}{RGB}{18,68,115}
\definecolor{darkgreen}{rgb}{0.3,0.6,0.3}

\DeclareCaptionFont{green}{darkgreen}
\algrenewcommand{\alglinenumber}[1]{\footnotesize\textrm{#1:}}

\hypersetup{letterpaper,bookmarksopen,bookmarksnumbered,
pdfpagemode=UseOutlines,
colorlinks=true,
linkcolor=blue,
anchorcolor=blue,
citecolor=blue,
filecolor=blue,
menucolor=blue,
urlcolor=blue,
pdfinfo={
Title={Effective Footstep Planning for Humanoids Using Homotopy-Class Guidance},
Author={Vinitha Ranganeni, Oren Salzman, Maxim Likhachev}},
pdfproducer={LaTeX},
pdfcreator={pdfLaTeX},
pdfsubject={Robotics; Motion Planning},
pdfkeywords={Footstep Planning; Heuristics; Humanoids; Homotopy Classes}
}

\newcounter{hypothesis}[section]

\newcounter{question}[section]
\newenvironment{question}[1][]
{ \refstepcounter{question}\par\smallskip
\textbf{Q.\thequestion} #1 \rmfamily}{\vspace{0.5ex}}

\newcounter{set}[section]
\newenvironment{set}[1][]
{ \refstepcounter{set}\par\vspace{0.3ex}
\textbf{S.\theset} #1 \rmfamily}{\vspace{0.3ex}}

\definecolor{planned}{HTML}{31a354}
\definecolor{random}{HTML}{636363}
\definecolor{learned}{HTML}{a6611a}
\definecolor{basic}{HTML}{756bb2}
\definecolor{pcrrt}{HTML}{3182bd}
\definecolor{brrt}{HTML}{e6550d}

\newcommand{\CaptionCircle}[1]{\tikz{\draw[#1,fill=#1] (0,0) circle (.5ex);}}

\makeatletter
\def\endthebibliography{%
  \def\@noitemerr{\@latex@warning{Empty `thebibliography' environment}}%
  \endlist
}
\makeatother

\newcommand{\figref}[1]{Fig.~\ref{#1}} 
\newcommand{\subfigref}[2]{\figref{#1}.\subref*{#2}} 
\newcommand{\aref}[1]{Alg.~\ref{#1}} 
\newcommand{\lineref}[1]{Line~\ref{#1}} 
\newcommand{\alglinesref}[3]{\aref{#1}, lines~\ref{#2}-\ref{#3}} 
\newcommand{\Linessref}[3]{Lines~\ref{#1}-\ref{#2},~\ref{#3}} 

\newcommand{\ques}[1]{\textbf{Q.\ref{#1}}} 
\newcommand{\hset}[1]{\textbf{S.\ref{#1}}}

\newcommand{\ignore}[1]{}

\def\P{\mathcal{P}}  
  
\def\O{\mathcal{O}}

 \def\W{\mathcal{W}} \def\R{\mathcal{R}}

\def\dR{\mathbb{R}}

\newcommand{\Cpp}{C\raise.08ex\hbox{\tt ++}\xspace}

\def\HiLi{\leavevmode\rlap{\noindent\hbox to 0.9\columnwidth{\color{yellow!50}\leaders\hrule height 0.8\baselineskip depth .8ex\hfill}}}

\newcommand\algname[1]{\textsf{#1}\xspace}
\newcommand\mhastar{\algname{MHA*}}

\newcommand\hbsp{\algname{HBSP}}
\newcommand\astar{\algname{A*}}

\newcommand{\qg}{\ensuremath{q_{\text{goal}}}\xspace}

\newcommand{\Cfoot}{\ensuremath{\mathcal{X}_{\text{footstep}}}\xspace}

\newcommand{\Chum}{\ensuremath{\mathcal{X}_{\text{humanoid}}}\xspace}

\newcommand{\Wproj}{\ensuremath{\mathcal{W}_{2}}\xspace}
\newcommand{\Wwork}{\ensuremath{\mathcal{W}_{3}}\xspace}
\newcommand{\Wmultproj}{\ensuremath{\mathbb{W}}\xspace}

\newcommand{\Vfoot}{\ensuremath{\mathcal{V}_{\Cfoot}}\xspace}
\newcommand{\Efoot}{\ensuremath{\mathcal{E}_{\Cfoot}}\xspace}
\newcommand{\Gproj}{\ensuremath{\mathcal{G}_{\Wproj}}\xspace}
\newcommand{\Gmultproj}{\ensuremath{\mathcal{G}_{\Wmultproj}}\xspace}
\newcommand{\Vproj}{\ensuremath{\mathcal{V}_{\Wproj}}\xspace}
\newcommand{\Eproj}{\ensuremath{\mathcal{E}_{\Wproj}}\xspace}

\newcommand{\VmultprojS}{\ensuremath{\mathcal{V}_{\Wmultproj}^h}\xspace}

\newcommand{\GmultprojS}{\ensuremath{\mathcal{G}_{\Wmultproj}^h}\xspace}
\newcommand{\GprojS}{\ensuremath{\mathcal{G}_{\Wproj}^h}\xspace}
\newcommand{\VprojS}{\ensuremath{\mathcal{V}_{\Wproj}^h}\xspace}
\newcommand{\EprojS}{\ensuremath{\mathcal{E}_{\Wproj}^h}\xspace}
\newcommand{\GfootS}{\ensuremath{\mathcal{G}_{\Cfoot}^h}\xspace}
\newcommand{\VfootS}{\ensuremath{\mathcal{V}_{\Cfoot}^h}\xspace}
\newcommand{\EfootS}{\ensuremath{\mathcal{E}_{\Cfoot}^h}\xspace}

\newcommand{\Mfoot}{\ensuremath{\mathcal{M}}\xspace}

\newcommand{\Hfoot}{\ensuremath{\mathcal{H}_{\text{footstep}}}\xspace}

\makeatletter
\def\thmhead@plain#1#2#3{%
  \thmname{#1}\thmnumber{\@ifnotempty{#1}{ }\@upn{#2}}%
  \thmnote{ {\the\thm@notefont#3}}}
\let\thmhead\thmhead@plain
\makeatother

\algrenewcommand\textproc{}

\begin{document}
%
\date{}
\title{Effective Footstep Planning Using Homotopy-Class Guidance}
\cortext[cor1]{Corresponding author}

\author[vr]{Vinitha Ranganeni\corref{cor1}}
\ead{vinitha@cs.uw.edu}
\author[sc]{Sahit Chintalapudi}
\ead{schintalapudi@gatech.edu}
\author[os]{Oren Salzman}
\ead{osalzman@andrew.cmu.edu}
\author[os]{Maxim Likhachev}
\ead{maxim@cs.cmu.edu}
\address[vr]{Paul G. Allen School of Computer Science \& Engineering, University of Washington, Seattle, WA, United States}
\address[sc]{College of Computing, Georgia Institute of Technology, Atlanta, GA, United States}
\address[os]{Robotics Institute, Carnegie Mellon University, Pittsburgh, PA, United States}

\begin{abstract}
Planning the motion for humanoid robots is a computationally-complex task due to the high dimensionality of the system. Thus, a common approach is to first plan in the low-dimensional space induced by the robot's feet---a task referred to as \emph{footstep planning}. This low-dimensional plan is then used to guide the full motion of the robot. One approach that has proven successful in footstep planning is using search-based planners such as \astar and its many variants. To do so, these search-based planners have to be endowed with effective heuristics to efficiently guide them through the search space. However, designing effective heuristics is a time-consuming task that requires the user to have good domain knowledge. Thus, our goal is to be able to effectively plan the footstep motions taken by a humanoid robot while obviating the burden on the user to carefully design local-minima free heuristics. To this end, we propose to use user-defined homotopy classes in the workspace that are intuitive to define. These homotopy classes are used to automatically generate heuristic functions that efficiently guide the footstep planner.
Additionally, we present an extension to homotopy classes such that they are applicable to complex multi-level environments.
We compare our approach for footstep planning with a standard approach that uses a heuristic common to footstep planning.
In simple scenarios, the performance of both algorithms is comparable.
However, in more complex scenarios our approach allows for a speedup in planning of several orders of magnitude when compared to the standard approach.
\end{abstract}

\maketitle

\section{Introduction}
\label{sec:intro}

\begin{figure}[!t]
    \captionsetup[subfigure]{position=top,textfont=scriptsize,singlelinecheck=off,justification=raggedleft,aboveskip=0pt,margin=0pt,labelfont={color=white}}

    \centering
    \vspace{-1.0em}
    \subfloat[]{
        \hspace{2.2em}
        \includegraphics[width=0.7\textwidth]{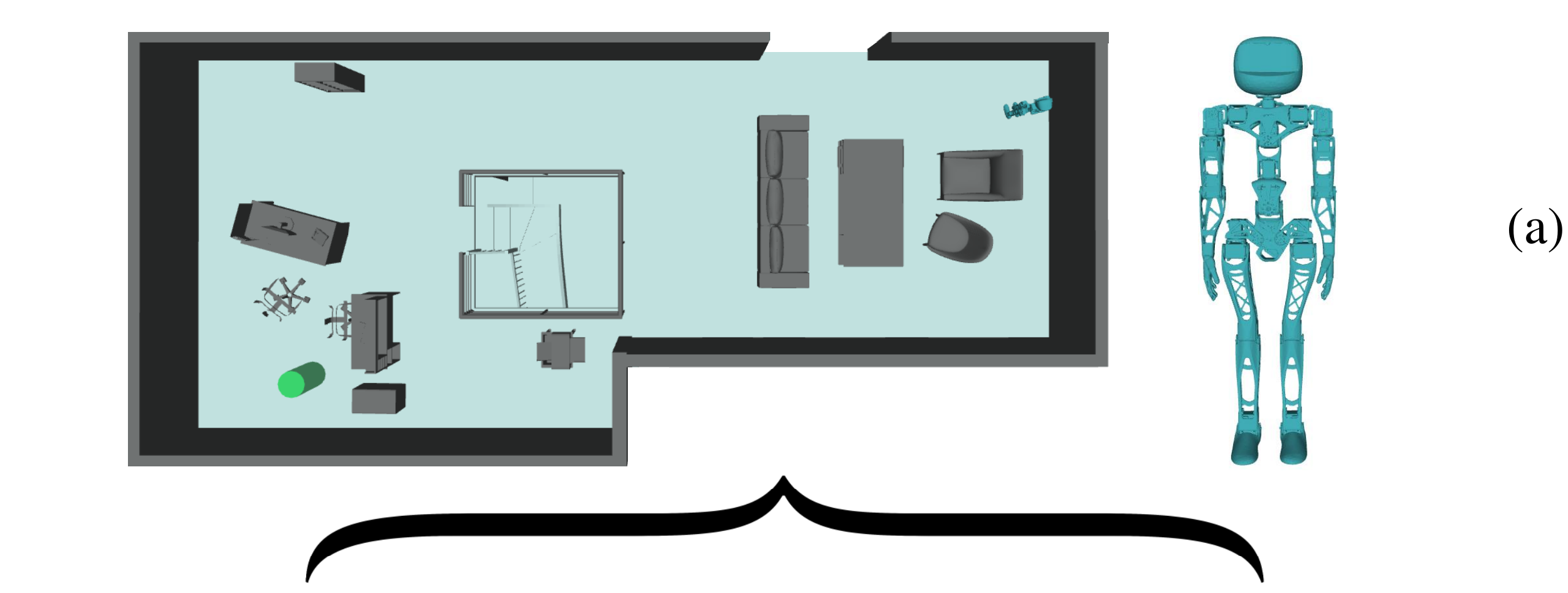}
        \label{fig:hbsp_split_1}
    }
    \vspace{-2.0em}
    \newline
    \subfloat[]{
        \hspace{2.2em}
        \includegraphics[width=0.7\textwidth]{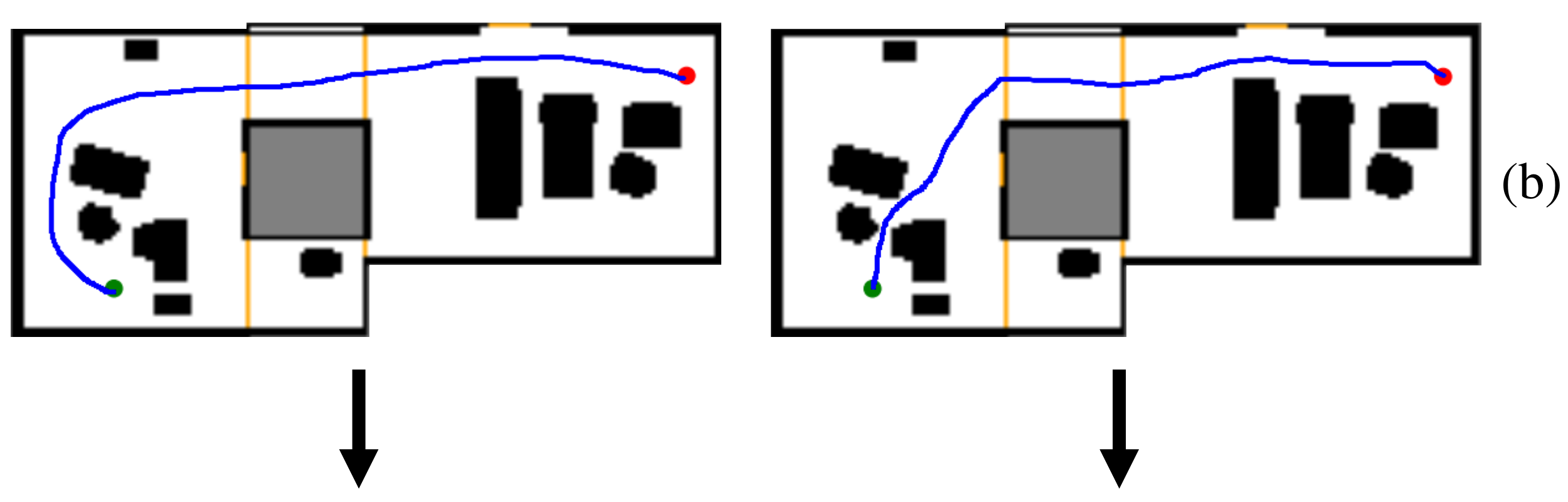}
        \label{fig:hbsp_split_2}
    }
    \vspace{-2.0em}
    \newline
    \subfloat[]{
        \hspace{2.2em}
        \includegraphics[width=0.7\textwidth]{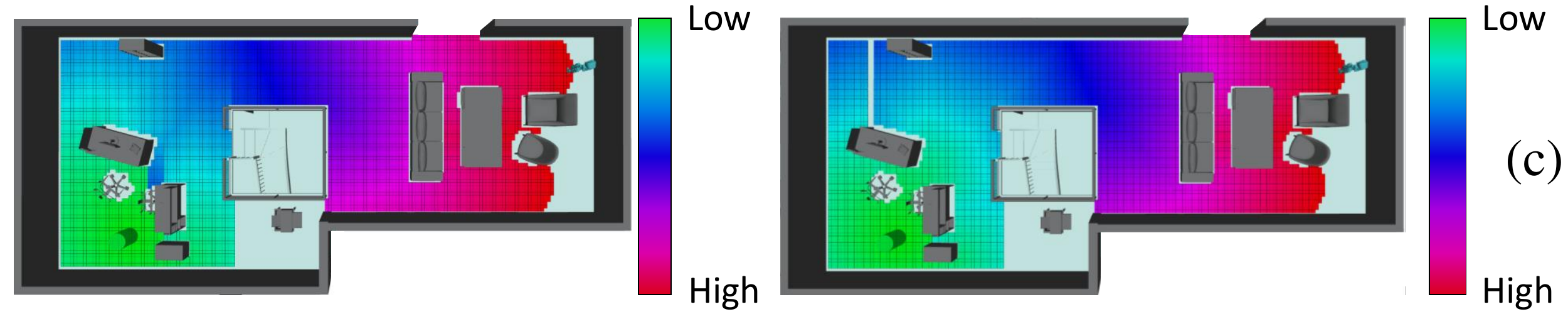}
        \label{fig:hbsp_split_3}
    }
    \vspace{-2.0em}
    \newline
    \subfloat[]{

        \includegraphics[width=0.7\textwidth]{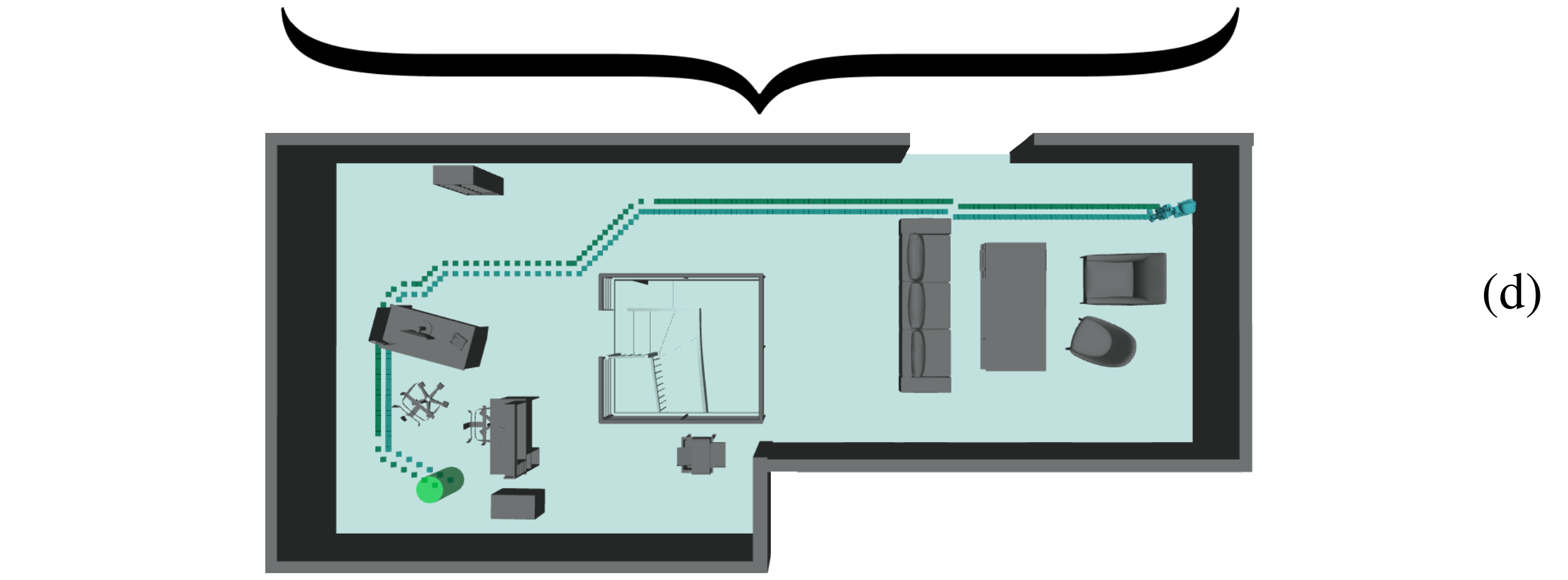}
        \label{fig:hbsp_split_4}
    }
    \caption{\protect\subref{fig:hbsp_split_1}~A Poppy humanoid robot~\cite{poppy} has to navigate to a goal region denoted by the green cylinder.~\protect\subref{fig:hbsp_split_2}~A user provides two reference paths in a 2D projection of the workspace.~\protect\subref{fig:hbsp_split_3}~A color map of the heuristic values for each homotopy-based heuristic. A single heuristic is constructed for each homotopy class of the reference paths.~\protect\subref{fig:hbsp_split_4}~The footstep planner uses both heuristics to quickly find a path to the goal region.}
    \label{fig:baseline_v_homotopic}
\end{figure}

Humanoid robots have been shown as an effective platform for performing a multitude of tasks in human-structured environments~\cite{kajita2014introduction}.
However, planning the motion of humanoid robots is a computationally-complex task due to the high dimensionality of the system.
Thus, a common approach to efficiently compute paths in this high-dimensional space is to guide the search using footstep motions which induce a lower-dimensional search space~\cite{GCBSL11,garimort11humanoid}. This lower-dimensional search space is an eight-dimensional configuration space of a humanoid robot's feet consisting of the $x$, $y$, $z$ positions and orientation for each foot.

One approach that has been successful in footstep planning is using search-based planners such as \astar~\cite{chestnutt05footstep,chestnutt07adaptive} and its anytime variants~\cite{garimort11humanoid,hornung12anytime}.
To plan footstep motions efficiently, these planners require effective heuristics (cost-to-go estimates) to guide the search.
Effective heuristic functions should avoid regions where the search ceases to progress towards the goal or when this progress is extremely slow (regions often referred to as a ``local minima'' or a ``depression regions''~\cite{I92}).

Consider, for example,~\subfigref{fig:baseline_v_homotopic}{fig:hbsp_split_1}.
Planning a path that passes between the desks is either unfeasible or may require expanding a massive number of states in order to precisely capture the sequence of configurations in which the robot does not collide  with obstacles.
The aforementioned challenges make constructing effective heuristics, that can intelligently reason about areas of the environment to avoid, a time consuming and tedious task that often requires considerable domain knowledge.

In this work we explore an alternative to manually designing heuristics.
Our key insight is that providing sketches of desirable trajectories, that can be represented as homotopy classes, can be used to automatically generate heuristic functions. As a side note, these homotopy classes may also be generated automatically, avoiding the need for sketches of desirable trajectories. However, this is out of the scope of our work. Instead, we present a method for generating heuristic functions given homotopy classes in 2D and Multi-level 2D\footnote{A Multi-level 2D environment is represented as a set of 2D workspaces.} environments.
Our method supports the use of multiple such heuristics to alleviate the requirement of capturing the different complexities induced by an environment in one single heuristic while maintaining guarantees on completeness.

Specifically, for each user-defined homotopy class in the workspace, we generate a heuristic function by running a search from the goal to the start configuration while restricting the search to expand only vertices within the specified homotopy-classes.
We call this algorithm Homotopy-Based Shortest Path, or \hbsp and detail it in Sec.~\ref{sec:approach}.
Additionally, we assume the existence of a simple-to-define heuristic which is admissible and consistent\footnote{A heuristic function is said to be admissible if it never overestimates the cost of reaching the goal.
A heuristic function is said to be consistent if its estimate is always less than or equal to the estimated distance from any neighboring vertex to the goal, plus the step cost of reaching that neighbor.}.

These heuristics are then used in Multi-Heuristic A* (\mhastar)~\cite{aine14mha,ASNHL16} which is a recently-proposed method that leverages information from multiple heuristics.
Roughly speaking, \mhastar simultaneously runs multiple \astar-like searches, one for each heuristic, and combines their different guiding powers in different stages of the search.
\mhastar~\cite{aine14mha,ASNHL16} is detailed in Sec.~\ref{sec:background} and the use of \mhastar~\cite{aine14mha,ASNHL16} together with \hbsp is detailed in Sec.~\ref{sec:approach}.

While our approach requires computing multiple heuristics before the planner can be executed, this can be done efficiently and thus takes a small fraction of the planning time.
Moreover, the extra computation invested in computing these heuristics allows to efficiently guide the footstep planner.
In some queries (Sec.~\ref{sec:experiments}), we present a speedup of several orders of magnitude when compared to standard approaches.

Compared to our prior conference publication~\cite{ROL18}, in this paper we extend the representation of homotopy classes, from planar 2D environments, to also be applicable in complex multi-level 2D environments and evaluate the effectiveness of the footstep planner in such domains.

\subsection{Motivating Example}
\label{sec:motivation}
Consider~\subfigref{fig:baseline_v_homotopic}{fig:hbsp_split_1} where a humanoid robot (shown in the upper right corner) has to navigate to the goal region denoted by the green cylinder.
Footstep planning for the humanoid plans in an 8D space defined by the $(x, y, z)$ position and orientation of each foot.
We calculate a simple heuristic by running a Dijkstra search backwards from the goal vertex $\qg$ to every vertex in the 2D workspace.
When executing the footstep planner with only this backward 2D Dijkstra heuristic $\mathcal{H}_{\text{Dijk}}$, the search is guided through the narrow passage between the desks, as it is the shortest path to the goal region. However, it is not feasible for the robot to pass through this region. After spending a significant amount of time expanding near the narrow passage, the heuristic eventually guides the search around the obstacles.

\hbsp, on the other hand, takes guidance from the user to determine which homotopy classes the heuristic functions should guide the search through.
In our example, it was unclear to the user whether the robot could pass through the narrow passage between the desks. Thus, the user provided two reference paths: (i) around the obstacles and (ii) through the narrow passage~(\subfigref{fig:baseline_v_homotopic}{fig:hbsp_split_2}) that were used to compute two corresponding heuristic functions~(\subfigref{fig:baseline_v_homotopic}{fig:hbsp_split_3}). While the path our footstep planner produced~(\subfigref{fig:baseline_v_homotopic}{fig:hbsp_split_4}), using these heuristics, is not the shortest path, there was approximately a \textbf{463 times} speedup in the planning time. It is also important to note that the performance of the planner is not hindered by poor quality heuristics. While one homotopy-based heuristic sought to guide the search through the narrow passage, another heuristic quickly guided the search around the obstacles.

\section{Related Work}
\label{sec:related_work}

In this section we describe related work on using search-based planning algorithms for footstep planning.
In Sec.~\ref{sec:footstep} we describe commonly-used search-based planning algorithms in the context of footstep planning. We also describe  previous work on dynamically generating heuristics.
In Sec.~\ref{sec:homotopy} we briefly mention how homotopy classes have been used in the broader context of motion planning.

\subsection{Footstep Planning Using Search-Based Planning}
\label{sec:footstep}
Several approaches for footstep planning have been proposed over the last few years.
Chestnutt \textit{et al.}  were the first to propose using \astar to plan around and over planar obstacles~\cite{chestnutt05footstep,chestnutt07adaptive}. However, the heuristic function used
was not consistent, and thus could suffer from long planning
time.

Indeed, as mentioned in Sec.~\ref{sec:intro} it can be  extremely difficult to hand
craft heuristics that can both efficiently guide the footstep
planner and maintain guarantees on the quality of the solution obtained.
Thus, one approach to speed up footstep
planning without manually designing informative heuristics
was to use anytime variants of \astar with simple-to-define
heuristics.
This was done using \algname{D* lite}~\cite{garimort11humanoid}
or using  \algname{ARA*} and \algname{R*}~\cite{hornung12anytime}

These anytime algorithms sacrifice optimality for speed.
An alternative approach to obtain efficient planners would be using informative heuristics.
One approach to obtain such informative heuristics is applying different learning methods~\cite{us13learning,arfaee11learning,thayer11learning,bhardwaj17learning}.
While highly effective, this approach requires a large amount of training data and a good feature set for training.
%

\subsection{Motion Planning Using Homotopy Classes}
\label{sec:homotopy}
Homotopy classes have been frequently used to model the motion of a robot tethered to a fixed base point~\cite{BLK12,GS98,SH15,BKHS15, igarashi2010homotopic}. The presence of obstacles introduces geometric and topological constrains for these robots. The constraints can create scenarios where the goal can only be reached if the cable configuration lies within a specific homotopy class, thereby making homotopy-based motion planning incredibly useful.

Homotopy classes have also been used in the context of human-robot interaction where a human wishes to restrict a robot's motion to specific homotopy classes~\cite{yi16homotopy}. Additionally, homotopy classes have been used to improve navigation for mobile robots~\cite{kuderer2013teaching, kuderer2014online, rosmann2015planning}.
For general approaches to explore and compute shortest paths in different homotopy classes, see~\cite{BLK12,B10,KBGK13}.

In our prior conference publication~\cite{ROL18}, we proposed a method for constructing heuristics based on user-provided homotopy classes in planar 2D environments. We used these heuristics to effeciently guide a footstep planner through complex environments and showed a speedup of several orders of magnitude compared to standard approaches. In this work, we extend the representation of homotopy-classes such that they are applicable in multi-level 2D environments.

\begin{figure*}[!t]
    \centering
    \subfloat[]{
        \includegraphics[width=0.24\textwidth]{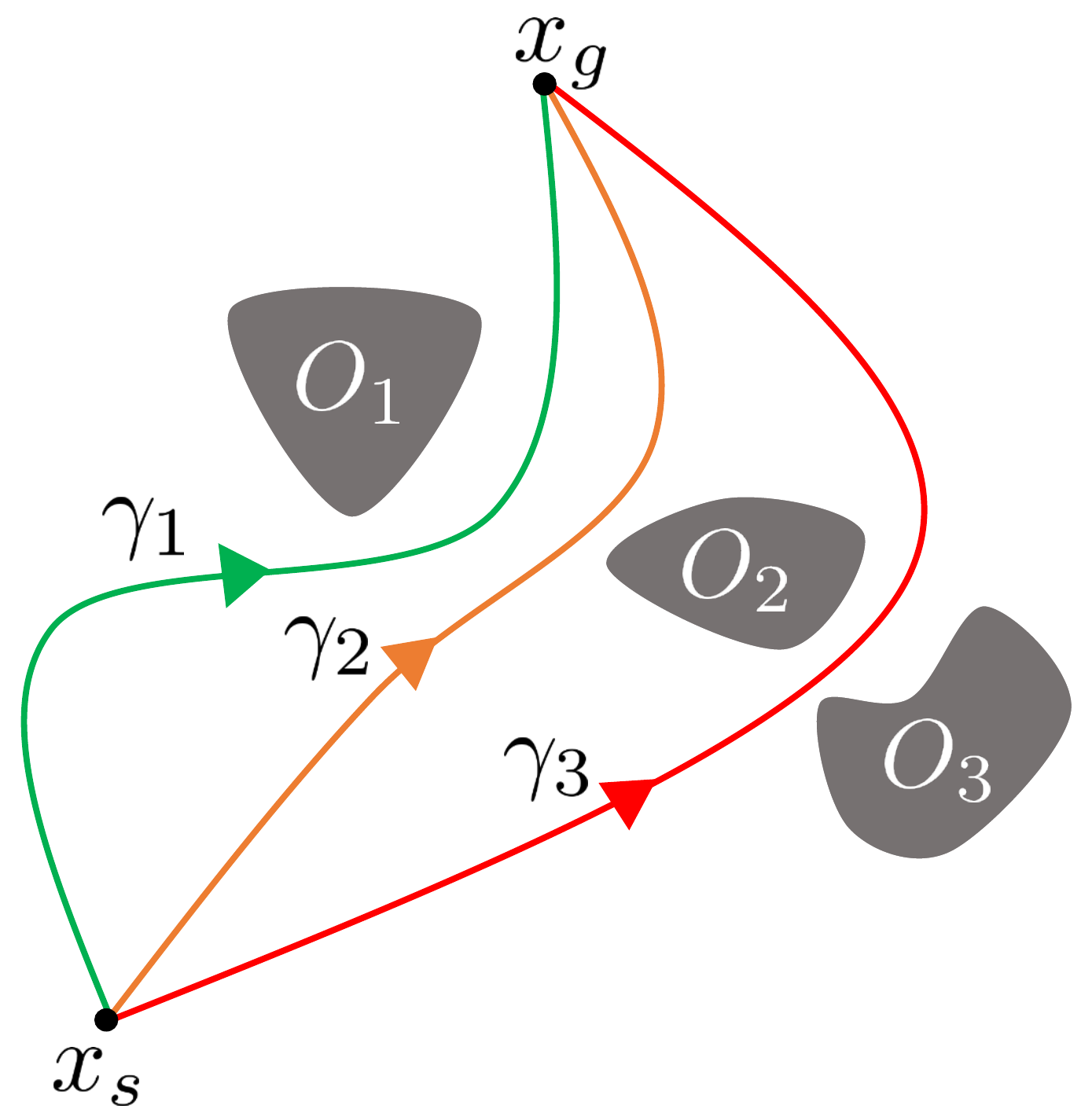}
        \label{fig:homotopy_1}
    }
    \subfloat[]{
        \includegraphics[width=0.44\textwidth]{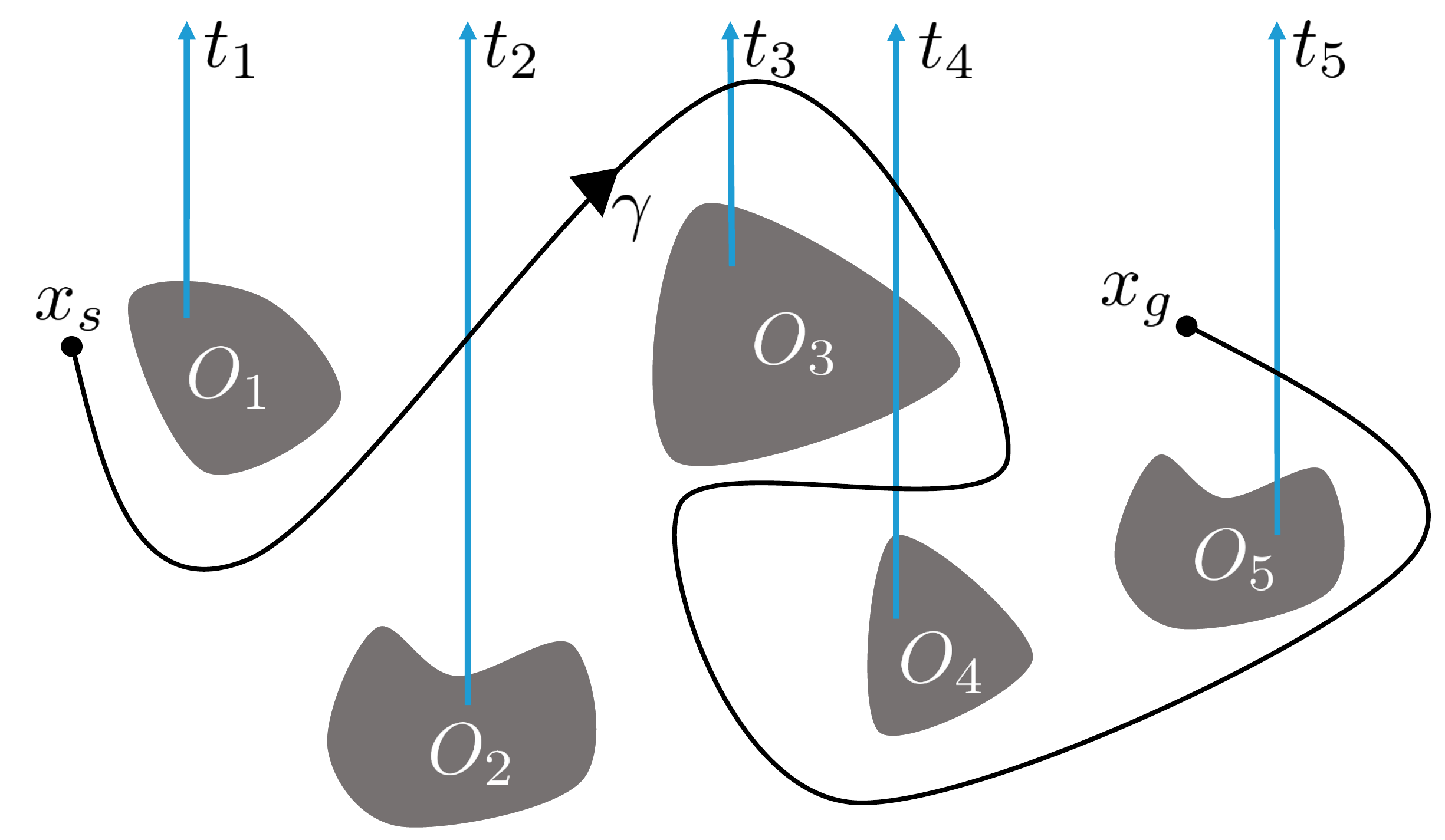}
        \label{fig:homotopy_2}
    }
    \subfloat[] {
        \includegraphics[width=0.26\textwidth]{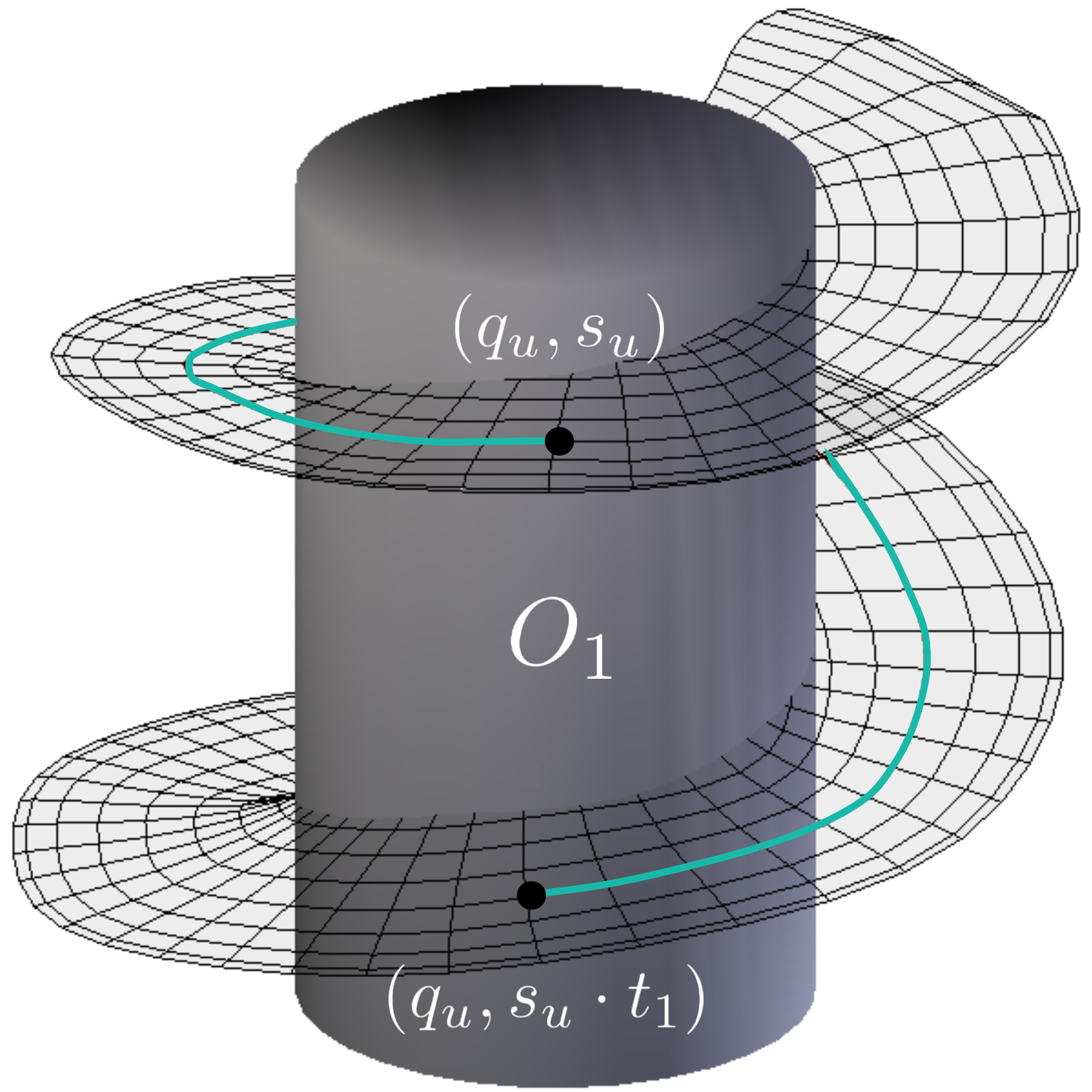}
        \label{fig:aug}
    }
    \label{fig:homotopy_classes}
     \caption{\protect\subref{fig:homotopy_1}~$\gamma_1$ and $\gamma_2$ are in the same homotopy class, however, $\gamma_3$ is in a different homotopy class because of obstacle~$O_2$. \protect\subref{fig:homotopy_2}~The signature for this curve is $t_2t_3t_4\bar{t_4}\bar{t_5}$. The homotopy invariant or \textit{h-signature} of curve $\gamma$ is $t_2t_3\bar{t_5}$. \protect\subref{fig:aug}~\GprojS has vertices $(q_u,s_u)$ and $(q_u,s_u \cdot t_1)$ that have the same configuration~$q_u$ but different signatures $s_u$ and $s_u \cdot t_1$.
     }
\end{figure*}

\section{Algorithmic Background}
\label{sec:background}
In this section we describe the algorithmic background necessary to understand our approach.
In Sec.~\ref{sec:homotopyClasses} we formally define the notion of homotopy classes and how to efficiently identify if two curves are in the same homotopy class---a procedure that will be used in our homotopy-based shortest-path algorithm (\hbsp).
In Sec.~\ref{sec:mha} we provide background on \mhastar~\cite{aine14mha,ASNHL16} which we will use to compute footstep plans by incorporating the heuristics computed by~\hbsp.

\subsection{Homotopy classes of curves}
\label{sec:homotopyClasses}
Informally, two continuous functions are called \emph{homotopic} if one can be ``continuously deformed'' into the other (See~\figref{fig:homotopy_1}).
If two curves are embedded in the plane, a straightforward characteristic exists to identifying and computing the homotopy class  of a curve~\cite{armstrong2013basic}.

In contrast, defining homotopy classes in three dimensions, in a manner that contains the same amount of information as homotopy classes in two dimensions, is very difficult. In 2D, any finite obstacle on a plane can induce multiple homotopy classes~\cite{B11}. However, homotopy classes in 3D can only be induced by obstacles with holes or handles, or with obstacles stretching to infinity in two directions~\cite{B11}. For example, a sphere does not induce any homotopy classes. A torus-shaped obstacle, on the other hand, will produce two types of homotopy classes: (1) For the trajectories passing through the hole of the torus and (2) for the trajectories passing outside the hole of the torus. Since defining homotopy classes in 3D is very difficult, we present a simple-yet-effective extension in Sec.~\ref{sssec:multiplehomotopyClasses} to define homotopy classes in multi-level 2D environments. This extension allows to define homotopy classes for more complex 3D environments that cannot be represented in 2D without placing restrictions on the type of obstacles.

\subsubsection{Homotopy classes in 2D environments}
\label{sssec:singlehomotopyClasses}

Let $\mathcal{W}_2 \subset \mathbb{R}^2$ be a subset of the plane (in our work this will be a two-dimensional projection of the three-dimensional workspace where the robot moves)
and let $\mathcal{O} = \{\mathcal{O}_1,...,\mathcal{O}_m\}$ be a set of obstacles
(in our work, these will be projections of the three-dimensional workspace obstacles).

In order to identify if two curves $\gamma_1, \gamma_2 \in \mathcal{W}_2~\backslash~\mathcal{O}$ that share the same endpoints are homotopic we use the notion of \textit{$h$-signature} (see~\cite{BLK12,GS98,SH15}).
The $h$-signature uniquely identifies the homotopy class of a curve.
That is, $\gamma_1$ and $\gamma_2$ have identical \textit{h}-signatures if and only if they are homotopic.

In order to define the $h$-signature, we choose a point $p_k \in \mathcal{O}_k$ in each obstacle such that no two points share the same $x$-coordinate.
We then extend a vertical ray or ``beam'' $b_k$ towards~$y = +\infty$ from $p_k$.
Finally, we associate a letter $t_k$ with beam $b_k$ (See~\figref{fig:homotopy_2}).

Now, given $\gamma$, let $b_{k_1}, \ldots, b_{k_m}$ be the sequence of $m$ beams crossed when tracing $\gamma$ from start to end.
The \textit{signature} of $\gamma$, denoted by $s(\gamma)$, is a sequence of $m$ letters.
If $\gamma$ is intersected by the beam $b_k$, by crossing it from left to right (right to left), then the $i$'th letter is $t_k$ ($\bar{t_k}$, respectively).
The reduced word, denoted by $r(s(\gamma))$, is constructed by eliminating a pair of consecutive letters in the form of $t_k\bar{t_k}$ or $\bar{t_k}t_k$.
The reduced word $r(s(\gamma))$ is a \textit{homotopy invariant} for curves with fixed endpoints. It will be denoted as $h(\gamma) = r(s(\gamma))$ and called the \textit{$h$-signature} of $\gamma$.

\begin{figure*}[t!]
    \centering
    \subfloat[]{
        \includegraphics[width=0.46\textwidth]{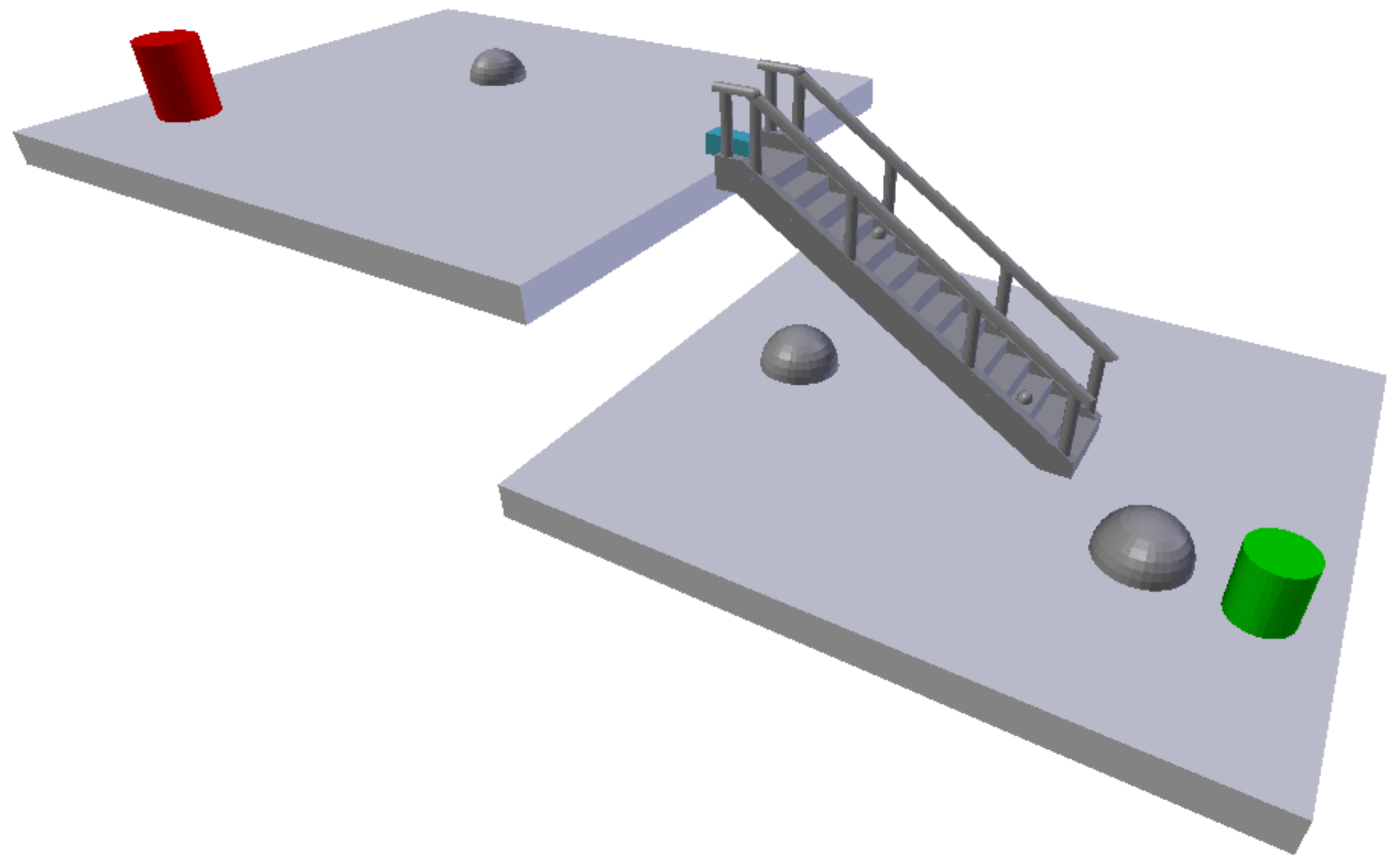}
        \label{fig:rampstart}
    }
    \subfloat[]{
        \includegraphics[width=0.46\textwidth]{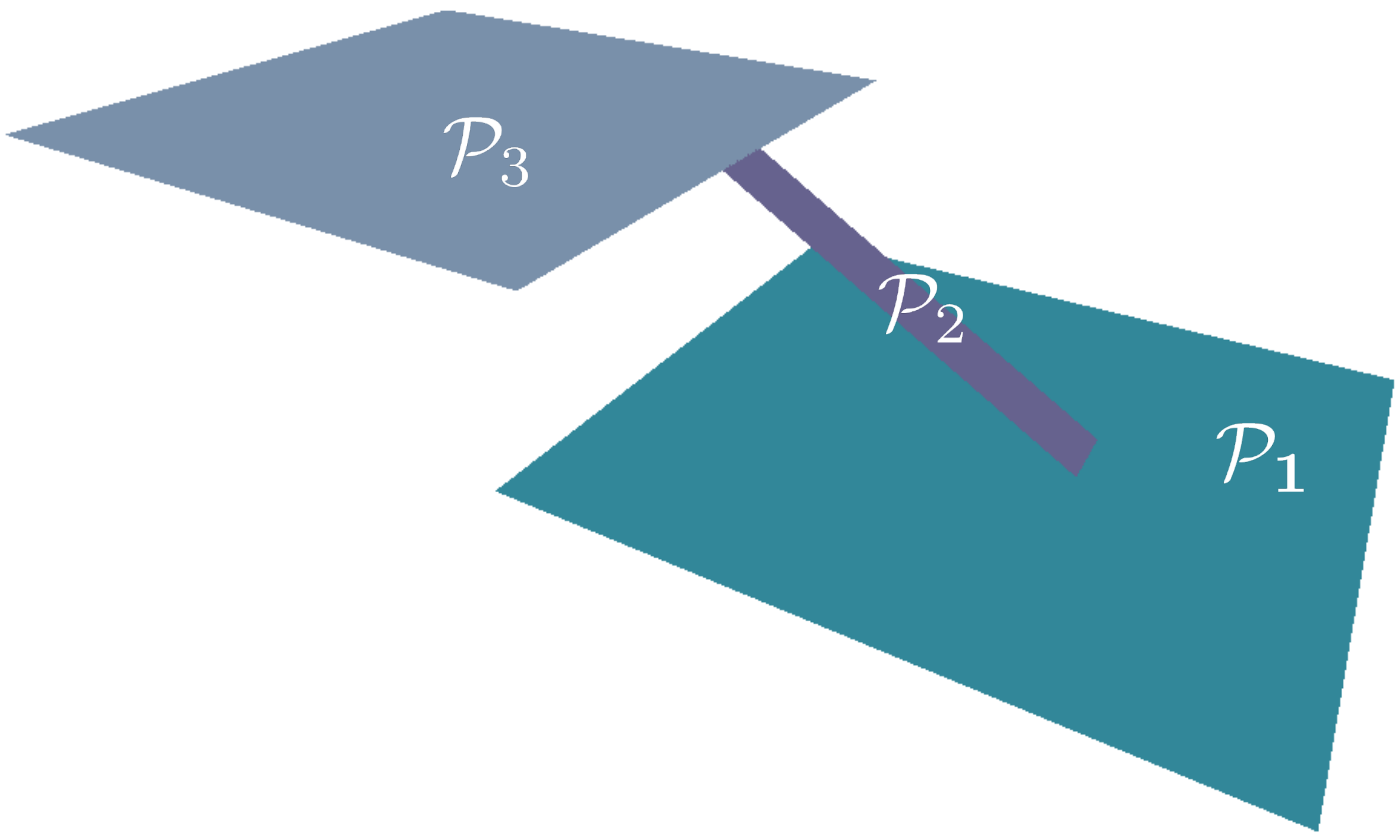}
        \label{fig:levels}
    }
    \newline
    \subfloat[]{
        \includegraphics[width=0.96\textwidth]{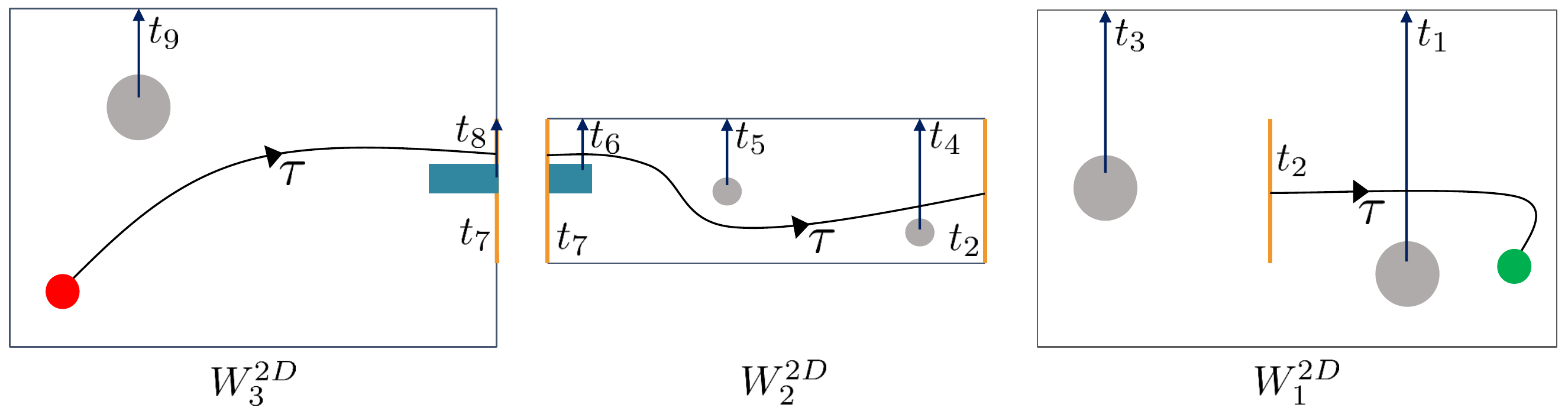}
        \label{fig:projection}
    }
    \caption{\protect\subref{fig:rampstart} A humanoid robot must navigate from the red cylinder to a goal region denoted by the green cylinder. \protect\subref{fig:levels} The set of surfaces $\mathbb{P} = \{\mathcal{P}_1, \mathcal{P}_2, \mathcal{P}_3\}$ are used to represent the world shown in \protect\subref{fig:rampstart}.~\protect\subref{fig:projection} $t_2$ and $t_7$ are the letters corresponding to gates $g_2$ and $g_7$ respectively (orange lines). The h-signature of curve $\tau$ is $t_8\bar{t_7}t_6t_4\bar{t_2}t_1$.}
    \vspace{-0.5em}
    \label{fig:guidanceui}
\end{figure*}

As we will see, our search algorithm~\hbsp will incrementally construct paths. After they are fully constructed, they will be in the same homotopy class as a given reference path.
Thus, it will be useful to understand how the $h$-signature of a curve $\gamma$, which is a concatenation of two curves $\gamma_1,\gamma_2$ can be easily constructed.
This reduced signature of $\gamma = \gamma_1 \cdot \gamma_2$ is simply the reduced signature of the concatenation of two curves' signatures $h(\gamma) = r(s(\gamma_1) \cdot s(\gamma_2))$.

\subsubsection{Homotopy classes in multi-level 2D environments}
\label{sssec:multiplehomotopyClasses}
Let $\W_3 \subset \mathbb{R}^3$ be our three-dimensional workspace. We assume that $\W_3$ can be represented as a set of bounded planes $\mathbb{P}$ called \textit{surfaces} (\figref{fig:levels}) (in our work these surfaces are manually created). Let $\mathcal{O} = \{\mathcal{O}_1,...,\mathcal{O}_m\}$ be a set of obstacles, $\O^i \subseteq \O$ be the obstacles that are within the xy-bounds of the $i$'th-surface $\P_i \in \mathbb{P}$. Additionally, we define a plane $P_{i \cap j} = a_{i \cap j}x + b_{i \cap j}y + c_{i \cap j}z + d_{i \cap j}$, bounded by the extents of the environment, that contains the interval $\P_i \cap \P_j$ and has a normal vector orthogonal to the z-axis.

We refine the set of obstacles as follows: as long as there exists some obstacle $\O_k$ such there is a point $p_k \in \O_k$ (where $p_k = (x_k, y_k, z_k)$), and there exists two surfaces $\P_i$ and $\P_j$ with $p_k \in \P_i \cap \P_j$, we subdivide $\O_k$ into two obstacles:
\begin{enumerate}
    \item $\{p_k \in \O_k~|~a_{i \cap j} \cdot x_k + b_{i \cap j} \cdot y_k + c_{i \cap j} \cdot z_k + d_{i \cap j} \leq 0\} \subset \O$, and
    \item $\{p_k \in \O_k~|~a_{i \cap j} \cdot x_k + b_{i \cap j} \cdot y_k + c_{i \cap j} \cdot z_k + d_{i \cap j} >~0\} \subset \O$.
\end{enumerate}
For example, in \figref{fig:rampstart} the blue obstacle spans over $\P_2$ and $\P_3$, thus it is divided into $\mathcal{O}_6 \subseteq \mathcal{O}^2$ and $\mathcal{O}_8 \subseteq \mathcal{O}^3$ (\figref{fig:projection}). We then project obstacles $\O^i$ onto $\P_i$ to generate a 2D workspace $\W^{2D}_{i} \subset \mathbb{R}^2$. The set of these 2D workspaces is $\mathbb{W}$ = $\{\W^{2D}_{1}, \W^{2D}_{2},...,\W^{2D}_{n}\}$. Additionally, we define a set of projected obstacles $\mathbb{O} =~\{\O^{2D}_{1}, \O^{2D}_{2},...,\O^{2D}_{m}\}$ and a set of projected obstacles $\mathbb{O}^i \subseteq \mathbb{O}$ that have been projected onto the \textit{i}'th workspace $\W^{2D}_{i}$.

To capture the \textit{h}-signature of a curve $\gamma$ that crosses multiple workspaces in $\mathbb{W}$, we introduce a variant of beams (see Sec.~\ref{sssec:singlehomotopyClasses}) called \textit{gates} that allow us to identify when a curve crosses from one workspace to another. A gate is a set of ``free points" (i.e. points that are not occupied by obstacles) formed by the intersection between two workspaces, $W^{2D}_{i} \in \mathbb{W}$ and $W^{2D}_{j} \in \mathbb{W}$. More specifically, the gate connecting $W^{2D}_{i}$ and $W^{2D}_{j}$ is denoted by
$$g_{k} = \{p \in W^{2D}_{i} \cap W^{2D}_{j}~\vert~\nexists \O \in \mathbb{O}^i \cup \mathbb{O}^j~\text{s.t.}~p\in O\}.$$
Note that there is only one gate connecting two workspaces, even if there are obstacles on the gate. When a trajectory crosses over such a gate, the gate letter (in the signature) will record the transition between workspaces and the obstacle letter(s) will record which section of the gate the trajectory is crossing over. Therefore, it is unnecessary to create multiple gates between the obstacles on $W^{2D}_{i} \cap W^{2D}_{j}$.
Furthermore $g_k$ cannot be empty; if an obstacle(s) completely overlap $W^{2D}_{i} \cap W^{2D}_{j}$, a gate does not exist at that intersection.

We can construct the signature of $\gamma$ using both gates and beams similar to the method described in~Sec.~\ref{sssec:singlehomotopyClasses}. Let $g_k$ be a gate that connects $W^{2D}_{i} \text{ to } W^{2D}_{j}$. If $i < j$ we add the letter $t_k$ to the signature. Otherwise, if $i > j$ we add $\bar{t_k}$ to the signature. While beams cannot intersect, beams and gates can intersect. We define an order for adding beams and gates when a trajectory $\gamma$ crosses them simultaneously. Let $b_q$ be a beam and $g_r$ be a gate connecting $W^{2D}_{i} \text{ to } W^{2D}_{j}$. If $\gamma$ crosses $b_q$ from left-to-right (right-to-left), $i < j$, and the point $p_q \in W^{2D}_{i}$, where $p_q$ is the point from which $b_q$ is extended, then $s(\gamma)~=~t_qt_r$ ($\bar{t_q}t_r$, respectively). Otherwise, if $p_q \in W^{2D}_{j} \setminus W^{2D}_{i} \cap W^{2D}_{j}$, then $s(\gamma)~=~\bar{t_r}t_q$ ($\bar{t_r}\bar{t_q}$, respectively). (\figref{fig:projection}). We must also maintain the beam and gate ordering when reducing words. Let $s(\gamma) = t_r \cdot t_q\bar{t_r}$. If $i < j$ and the point $p_q \in W^{2D}_{i}$ then $r(s(\gamma)) = t_q$.


\subsection{Multi-Heuristic A* (\mhastar)}
\label{sec:mha}
The performance of heuristic search-based planners, such as~\astar, depends heavily on the quality of the heuristic function. 
For many domains, it is difficult to produce a single heuristic function that captures all the complexities of the environment.
Furthermore, it is difficult to produce an admissible heuristic
which is a necessary condition for providing guarantees on solution quality and completeness.

One approach to cope with these challenges is by using \emph{multiple} heuristic functions.
\mhastar~\cite{aine14mha,ASNHL16} is one such approach that takes in a single admissible heuristic called the \emph{anchor} heuristic, as well as  multiple (possibly) inadmissible heuristics.
It then simultaneously runs multiple \astar-like searches, one associated with each heuristic, which allows to automatically combine the guiding powers of the different heuristics in different stages of the search.

Aine \textit{et al.}~\cite{aine14mha,ASNHL16} describe two variants of \mhastar, Independent Multi-Heuristic A* (\algname{IMHA*}) and Shared Multi-Heuristic A* (\algname{SMHA*}). Both of these variants guarantee completeness and provide bounds on sub-optimality.
In \algname{IMHA*} each individual search runs independently of the other searches
while in \algname{SMHA*}, the best path to reach each state in the search space is shared among all searches.
This allows each individual search to benefit from progress made by other searches.
This also allows \algname{SMHA*} to combine partial paths found by different searches which, in many cases, makes \algname{SMHA*} more powerful than \algname{IMHA*}.
Therefore in this work we will use \algname{SMHA*}. For brevity we will refer to \algname{SMHA*} as \mhastar.

\section{Homotopy-Based Planning}
\label{sec:approach}

Our footstep planner is comprised of \hbsp, which generates the heuristic functions from a set of user-defined homotopy classes, and \mhastar~\cite{aine14mha,ASNHL16}, which simultaneously uses these heuristics to find a feasible path.
We begin by defining a taxonomy of the different spaces we consider (Sec.~\ref{sec:tax}).
We then detail our motion planner (Sec.~\ref{sec:alg}) and \hbsp  (Sec.~\ref{sec:hbsp}).

\subsection{Taxonomy of Search Spaces}
\label{sec:tax}
\subsubsection{Search spaces}

Let $\Wwork \subset \mathbb{R}^3$ be the three-dimensional workspace in which the robot operates and $\Wproj \subset \mathbb{R}^2$ be its two-dimensional projection. Here, we \textit{pessimistically project} our 3D environment into a 2D workspace. This approach projects a cell~$(x, y, z) \in W_3$ as an obstacle cell $(x, y) \in W_2$ if there is at least one $z$-value for which $(x, y, z)$ is occupied. However, this approach does not work in complex multi-leveled terrains as a pessimistic projection would no longer provide useful information. In order to represent such environments we use a set of 2D workspaces \Wmultproj.

Let \Chum, \Cfoot  be the configuration spaces of the humanoid robot\footnote{Planning in \Chum  is out of the scope of this paper. We mention this space to provide the reader with a complete picture of all the search spaces relevant to planning the motion of a humanoid robot.} and the humanoid's footsteps respectively. Specifically,~\Chum is high-dimensional (over several dozens of dimensions) space, while~\Cfoot is a (relatively) low-dimensional space. In this work~\Cfoot is the eight-dimensional space $SE(3) \times SE(3)$ denoting the position $(x, y, z)$ and orientation of each of the robot's feet. Let $\Mfoot: \Cfoot \rightarrow W_{i}^{2D}$ be a mapping projecting footstep configurations to the corresponding workspaces projection $W_{i}^{2D} \subset \Wmultproj$. We use the mapping
$$\Mfoot(x_1, y_1, z_1, \theta_1, x_2, y_2, z_2, \theta_2) = \left((x_1 + x_2) / 2 , (y_1 + y_2) / 2 \right).$$
\indent Finally, we assume that each space~$\mathcal{X}$ induces a graph
$\mathcal{G}_\mathcal{X} = \left(\mathcal{V}_\mathcal{X} ,\mathcal{E}_\mathcal{X}  \right)$ embedded in $\mathcal{X}$.
For example, the vertices can be defined by overlaying the space~$\mathcal{X}$ with a grid and edges connecting every two nearby vertices.


\subsubsection{Augmented graphs}
In this work, we use homotopy classes to guide our footstep planner.
Thus, we will use the notion of \emph{augmented graphs} which, for a given graph $\mathcal{G}$ and a goal vertex $u_{\text{goal}}$, capture the different homotopy classes to reach every vertex in $\mathcal{G}$ from $u_{\text{goal}}$.
To define the augmented graphs, we first need to define the \emph{signature set}~$S(O,~\mathbb{G})$ of a set of obstacles $O$ and the set of gates $\mathbb{G}$.
Let $B(O)$ be the beams associated with the obstacles in $O$.
The signature set is defined as all the different $h$-signatures that can be constructed using~$B(O)$ and $\mathbb{G}$.
Note that $S(O,~\mathbb{G})$ is a countably infinite set.

The first graph we will augment is \Gproj. Let \Wproj denote any projected workspace.
Let $\GprojS = \left( \VprojS , \EprojS\right)$  denote this augmented graph induced by the projected workspace~\Wproj.
The set of vertices is defined as
$\VprojS  = \Vproj \times S(O,~\mathbb{G})$.
Namely, it consists of all pairs $(q, s)$ where $q$ is a vertex in $\Vproj$ and $s \in S(O,~\mathbb{G})$ is a signature.
The set of edges is defined as
\begin{align*}
\EprojS
 =  \{
    \left( (q_u,s_u), (q_v, s_v) \right)
    \vert
    (q_u,q_v) \in \Eproj \text{ and }
    h(s_u \cdot s_{u,v}^{\Wproj}) = h(s_v) \}.
\end{align*}
Here $s_{u,v}^{\Wproj}$ is the signature of the trajectory in \Wproj associated with the edge $(q_u,q_v)$.
Namely, \EprojS consists of all edges $(u,v)$ connecting vertices such that
(i)~there is an edge in~$\mathcal{E}_{\mathcal{W}_2}$ between the $q_u$ and $q_v$
and
(ii)~the reduced signature obtained by concatenating $s_u$ with the signature of the trajectory associated with the edge $(q_u,q_v)$ yields $s_v$.
It is important to note that \GprojS can have vertices that have the same configuration $q \in \Vproj$ but with different signatures (Fig.~\ref{fig:aug}).

Next, we augment the graph $\mathcal{G}_{\mathbb{W}}$. Let $\mathcal{G}_{\mathbb{W}}^h = (\mathcal{V}_{\mathbb{W}}^h, \mathcal{E}_{\mathbb{W}}^h)$ denote the augmented graph induced by a set of projected workspaces $\mathbb{W}$. Every projected workspace $\mathcal{W}_{i}^{2D} \subseteq \mathbb{W}$ has a corresponding augmented graph $\mathcal{G}_{\mathcal{W}_{i}^{2D}}^h = (V_{\mathcal{W}_{i}^{2D}}^h, E_{\mathcal{W}_{i}^{2D}}^h)$. Then the vertex set is given by $V_{\mathbb{W}}^h = \bigcup_{i = 1}^n V_{\mathcal{W}_{i}^{2D}}^h$ and the edge set is given by $E_{\mathbb{W}}^h = \bigcup_{i = 1}^n E_{\W_{i}^{2D}}^h$. Similar to \GprojS, \GmultprojS can have vertices that have the same configuration $q \in \VmultprojS$ but with different signatures.

\begin{figure}[t!]
    \centering
    \includegraphics[width=0.7\textwidth]{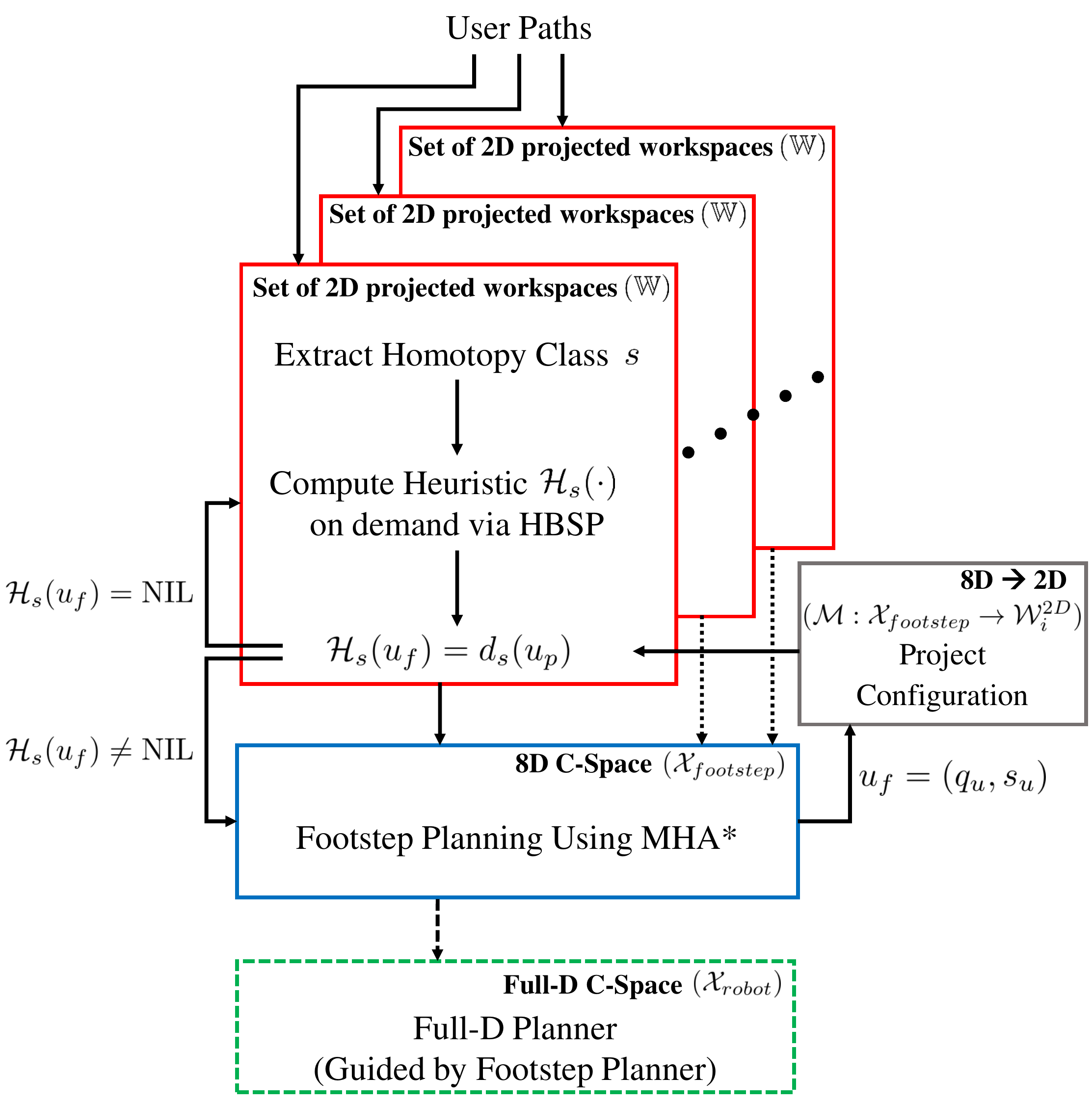}
    \caption{Overview of algorithmic approach for footstep planning.}
    \label{fig:alg_overview}
\end{figure}

Finally, the augmented graph induced by \Cfoot is denoted by
$\GfootS = \left( \VfootS , \EfootS\right)$.
Here,
the set of vertices is defined as
$\VfootS = \Vfoot \times S(O,~\mathbb{G})$
which is analogous to the definition of~$\VprojS$.
The set of edges is slightly more complex because \GfootS is not embedded in the plane.
Specifically, \EfootS is defined as
\begin{align*}
\EfootS
 =  \{
    \left( (q_u,s_u), (q_v, s_v) \right)
    \vert
    (q_u,q_v) \in \Efoot \text{ and }
    h(s_u \cdot s_{u,v}^{\Cfoot}) = h(s_v) \}.
\end{align*}

Here $s_{u,v}^{\Cfoot}$ is the signature of the trajectory in \Wproj associated with the projection of the edge $(q_u,q_v)$.
Namely let $\tilde{q}_u = \mathcal{M}(q_u)$ and $\tilde{q}_v = \mathcal{M}(q_v)$ be the projections of the two vertices $q_u$ and $q_v$, respectively and similarly let $(\tilde{q}_u,\tilde{q}_v)$ denote the projection of the trajectory  associated with edge $(q_u,q_v)$. Signature $s_{u,v}^{\Cfoot}$ is simply the signature of $(\tilde{q}_u,\tilde{q}_v)$.

\subsubsection{Heuristic functions}
A heuristic function for the footstep planner is a mapping
$
\Hfoot : \VfootS \rightarrow \dR_{\geq 0}.
$
In this work, we use the set of projected workspaces $\mathbb{W}$ to guide the footstep planner. We use $\mathbb{W}$ with no homotopy-based information to obtain a simple heuristic $\mathcal{H}_{\text{Dijk}}$ as well as a set of heuristics defined using a set of homotopy classes. The heuristic $\mathcal{H}_{\text{Dijk}}$ is obtained by running a Dijkstra search from the projection of the goal configuration to every vertex in $\Gmultproj$. This heuristic is often referred to as a ``backward 2D Dijkstra heuristic''\footnote{This heuristic is generally used in a single projected workspace $\Wproj$. We altered the heuristic such that it can be used in $\mathbb{W}$.}.

Our homotopy-based heuristics are defined using mappings $d_s: \VmultprojS \rightarrow \dR_{\geq 0}$
(one for each signature $s \in S$).
Each mapping  associates distances with augmented vertices of the projected workspace.
Here, a heuristic function  will be defined as
$\mathcal{H}_s (q_u,s_u) =
 d_s \left( \mathcal{M}(q_u), s_u \right) $
 and~$d_s(u)$ is the shortest path to reach the goal from vertex~$u$ by following the homotopy path defined by $s$.

\subsection{Algorithmic approach}
\label{sec:alg}

\begin{algorithm}[t!]
    \begin{algorithmic}[1]
            \Function{\tt{HBSP}}{$Q$, $\Gmultproj$, $q_g$, $u$, $S$}
            \Comment{$u = (q_u,s_u)$}
                \If{ $Q = \emptyset$ and $dist[(q_g,\wedge)]=$ NIL}
                    \State $dist[(q_g,\wedge)]\gets 0$
                    \State $Q$.\tt{add\_with\_priority}$\left( (q_g, \wedge),0 \right)$
                \EndIf
                \While{$Q \neq \emptyset$}
                    \State $v \gets$ \tt{$Q$.extract\_min}()\Comment{$v = (q_v,{s_v})$}
                     \tikzmk{A}{
                        \If{$v = u$}\label{alg:line:hbsp:goal_reached}
                            \State {\Return $(Q, dist[u])$}\label{alg:line:hbsp:return1}
                        \EndIf
                    \State $V_{\text{succ}} \gets$ \tt{succ($v$,$S$,$\Gproj$)}}
                    \tikzmk{B}
                    \boxit{cyan}\label{alg:line:hbsp:succ}

                    \For{$v' \in V_{\text{succ}}$}\Comment{$v' = (q'_v,s_v)$}
                        \State $alt \gets dist[v]$ + length($v$,$v'$)
                        \If{$dist[v'] = $ NIL }
                            \State $dist[v']\gets alt$
                            \State $Q$.\tt{add\_with\_priority}($v',dist[v']$)
                        \ElsIf{$alt < dist[v']$}
                            \State $dist[v'] \gets alt$
                            \State \parbox[t]{\dimexpr\linewidth-\algorithmicindent}{$Q$.\tt{decrease\_priority}($v,dist[v']$)\strut}
                        \EndIf
                    \EndFor
                \EndWhile
                \tikzmk{A}{
                \State \Return $(Q, \infty)$\label{alg:line:hbsp:return2}
            \EndFunction}
                    \tikzmk{B}
                    \boxitone{cyan}
    \end{algorithmic}
    \caption{Homotopy-Based Shortest Path Algorithm}
    \label{alg:hbsp}
\end{algorithm}

In this section we describe our algorithmic approach which is  depicted in Fig.~\ref{fig:alg_overview}.
The algorithm starts by obtaining a set of user-defined homotopy classes in the projected workspace~\Wproj using an intuitive graphical user interface (see Fig.~\ref{fig:gui}).
Each homotopy class is represented by a signature $s$. Let $S$ represent the set of all such signatures.
Recall that each signature $s \in S$ is used to compute a heuristic $\mathcal{H}_s$ by computing a distance function~$d_s$ using our Homotopy-Based Shortest Path (\hbsp) algorithm (Fig.~\ref{fig:alg_overview}, red boxes).

Our planner (Fig.~\ref{fig:alg_overview}, blue box) runs a \mhastar-search over~\cite{aine14mha,ASNHL16} the augmented graph \GfootS guided by the set of heuristics $\{\mathcal{H}_s \vert s \in S \}$ as well as the anchor heuristic~$\mathcal{H}_{\text{Dijk}}$. Since $\{\mathcal{H}_s \vert s \in S \}$ is constructed based on user-defined homotopy classes, it is possible for the user to provide homotopy classes such that every heuristic biases the search towards regions where there is no feasible path. However, since we are using \mhastar~\cite{aine14mha,ASNHL16}, the algorithm is complete and the user cannot prevent a path from being found. Furthermore, since our anchor heuristic is admissible, we maintain guarantees on sub-optimality.

It is important to note that the distance functions $\{d_s \vert s \in S \}$ are \emph{not} pre-computed for every vertex $u \in \VmultprojS$ by \hbsp (this would be unfeasible as \VmultprojS is not a finite set).
Instead, they are computed in an on-demand fashion.
Specifically, given a vertex $u \in \VmultprojS$ and a signature~$s$, \hbsp checks if $d_s(u)$ has been computed.
If it has, the value is returned and if not the algorithm continues to run a search from its previous state until $d_s(u)$ has been computed.
This process is depicted in Fig.~\ref{fig:alg_overview} by the two different edges leaving the red box to its left.

\subsection{Homotopy-Based Shortest Path}
\label{sec:hbsp}

\begin{algorithm}[t!]
  \begin{algorithmic}[1]
    \Function{\tt{succ}}{$u$, $S$, $\Gmultproj$}\Comment{$u = (q_u,{s_u})$}
        \State $V_{\text{nbr}} \gets$ \tt{neighbors}($u$)

            \If{$S \neq \emptyset$}
                \State {$\mathbb{S} \gets$ \tt{suffixes}($S$)\label{alg:line:succ:suffixes}}
                \For{$v \in V_{\text{nbr}}$}
                \Comment{$v = (q_v,{s}_{u,v})$}
                    \If{\tt{valid}($v$)}
                    \tikzmk{A}{
                        \If{$h({s_u} \cdot {s}_{u,v}) \notin \mathbb{S}$} \label{alg:line:succ:sig_restricted}
                            \State {$V_{\text{nbr}}$.remove($v$)\label{alg:line:succ:remove_v}}
                        \Else
                            \State {$v = (q_v, {s_u} \cdot {s}_{u,v})$\label{alg:line:succ:update_sig}}
                        \EndIf} \tikzmk{B} \boxitfour{cyan}
                    \Else
                        \State {$V_{\text{nbr}}$.remove($v$)}
                    \EndIf
                \EndFor
            \EndIf
        \tikzmk{A}{
        \State \Return $V_{\text{nbr}}$
        \EndFunction}
        \tikzmk{B}
        \boxitthree{cyan}
  \end{algorithmic}
  \caption{HBSP Successor Function}
  \label{alg:succ}
\end{algorithm}

Given a goal configuration~\qg and the graph~\Gmultproj,~\hbsp incrementally constructs the augmented graph~\GmultprojS by running a variant of Dijkstra's algorithm from the vertex $(\qg, \wedge)$.
Here, $\wedge$ denotes the empty signature.
For all vertices in $\VmultprojS$ that were constructed, the algorithm maintains a map $dist : \VmultprojS \rightarrow \R_{\geq 0}$ which captures the cost of the shortest path to reach vertices in $\VmultprojS$  from the vertex $(\qg, \wedge)$.
Given a vertex $(q_u, s_u) \in \GmultprojS$ and some user-defined signature $s$, this map $dist$ is used to compute the mapping~$d_s$ (which, in turn, is required to compute the heuristic function~$\mathcal{H}_s$).
Specifically,
\begin{align*}
d_s(q_u, s_u) = dist [q_u, h(s \cdot s_u)].
\end{align*}

Note that $s$ corresponds to a signature of the path defined from the vertex $(\qg,\wedge)$ towards the vertex $(q_{\text{start}},s)$, where $q_{\text{start}}$ is a projection of the robot's start configuration. However, $s_u$ is computed by the search as it progresses from $(q_{\text{start}},\wedge)$ to $(\qg,s)$. Therefore $h(s \cdot s_u)$ corresponds to the remaining portion of the homotopy-based path specified by $s$ after we remove its prefix that corresponds to $s_u$. For example, let $s=\bar{t_3}\bar{t_2}\bar{t_1}$ and $s_u = t_1t_2$, then $h(s \cdot s_u) = h(\bar{t_3}\bar{t_2}\bar{t_1}t_1t_2) = \bar{t_3}$. Here, $\bar{t_3}$ is the signature of remaining portion of the path to the goal specified by $s$.

As the  graph~\GmultprojS contains an infinite number of vertices, two immediate questions come to mind regarding this Dijkstra's-like search:
\begin{enumerate}
    \item []\begin{question} When should the search be terminated? \label{ques:q1} \end{question}
    \item []\begin{question} Should the search attempt to explore all of~\GmultprojS? \label{ques:q2} \end{question}
\end{enumerate}

We address \ques{ques:q1} by only executing the search if it is queried for a value of~$d_s$ which has not been computed.
Thus, the algorithm is also given a vertex $u$ and runs the search until~$d_s(u)$ is computed or until $d_s(u) > w_2 \cdot \mathcal{H}_{\text{Dijk}}(u)$. Here, $w_2$ is the search prioritization factor used in \mhastar~\cite{aine14mha,ASNHL16}.
This approach turns \hbsp to an online algorithm that produces results in a just-in-time fashion.
It is important to note that when the search is terminated, its current state (namely its priority queue) is stored. When the algorithm continues its search, it is simply done from the last state encountered before it was previously terminated.

We address \ques{ques:q2} when computing the successors of a vertex described in~\aref{alg:succ} by restricting the vertices we expand. During the search, when we expand a vertex $u \in \VmultprojS$ in our Dijkstra-like search, we prune away all its neighbors $v \in \VmultprojS$ that have invalid signatures $h(s_u \cdot s_{u,v})$~(\alglinesref{alg:succ}{alg:line:succ:sig_restricted}{alg:line:succ:update_sig}).
We define a valid signature $h(s_u \cdot s_{u,v})$ as one that is a \textit{suffix} of a signature $s \in S$. Let $\mathbb{S}$ be the collection of all such signatures. That is, any signature $h(s_u \cdot s_{u,v})$, such that $h(s)$ could potentially be reached as the search progresses.
More specifically, these suffixes identify the order in which certain beams can be crossed to reach a signature $h(s)$, where $s \in S$. For example, in~\figref{fig:homotopy_2}, $S = \{t_2t_3t_4\bar{t_4}\bar{t_5}\}$ and $\mathbb{S}$ of~$S$ is $\{t_2t_3\bar{t_5}, t_2t_3t_4, t_2t_3, t_2, \wedge\}$. Here, $t_1 \notin \mathbb{S}$ as this beam does not need to be crossed to reach $t_2t_3\bar{t_5}$. Additionally, $t_3t_2 \notin \mathbb{S}$ as the beams need to be crossed in the opposite order to obtain the signature $t_2t_3\bar{t_5}$. Note that $S$ is the set of un-reduced signatures to ensure that \hbsp is complete. If $S = \{t_2t_3\bar{t_5}\}$ (i.e. a set of \textit{h}-signatures), then $\mathbb{S} = \{t_2t_3\bar{t_5}, t_2t_3, t_2, \wedge\}$. In this scenario \hbsp would not reach the goal because it needs to expand vertices with the signature $t_2t_3t_4$ in order to obtain the signature $t_2t_3\bar{t_5}$.

The high-level description of our algorithm is captured in Alg.~\ref{alg:hbsp}.
The algorithm is identical to Dijkstra's algorithm\footnote{The only places where \hbsp differs from Dijkstra's algorithm are the lines highlighted in blue.} except that
(i)~when the cost $dist[u]$ is returned, the priority queue $Q$ is also returned~(\Linessref{alg:line:hbsp:goal_reached}{alg:line:hbsp:return1}{alg:line:hbsp:return2})
and
(ii)~the way the successors of an edge are computed~(\lineref{alg:line:hbsp:succ}).
Returning the queue $Q$ allows the algorithm to be called in the future with~$Q$ in order to continue the search from the same state.

\begin{figure}[!t]
    \centering
    \subfloat[]{
        \includegraphics[width=0.485\columnwidth]{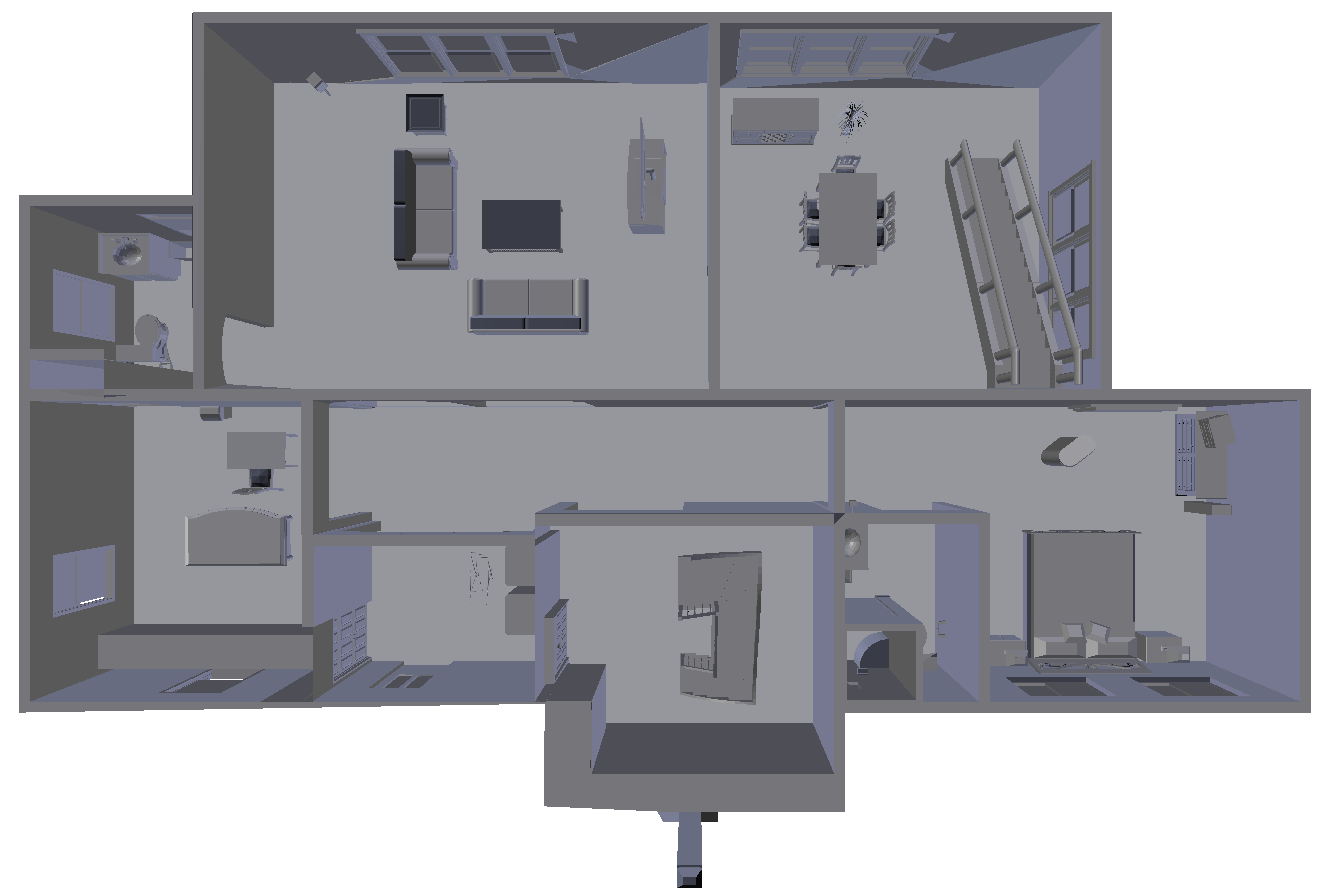}
        \label{fig:bottom_floor}
    }
    \subfloat[]{
        \includegraphics[width=0.515\columnwidth]{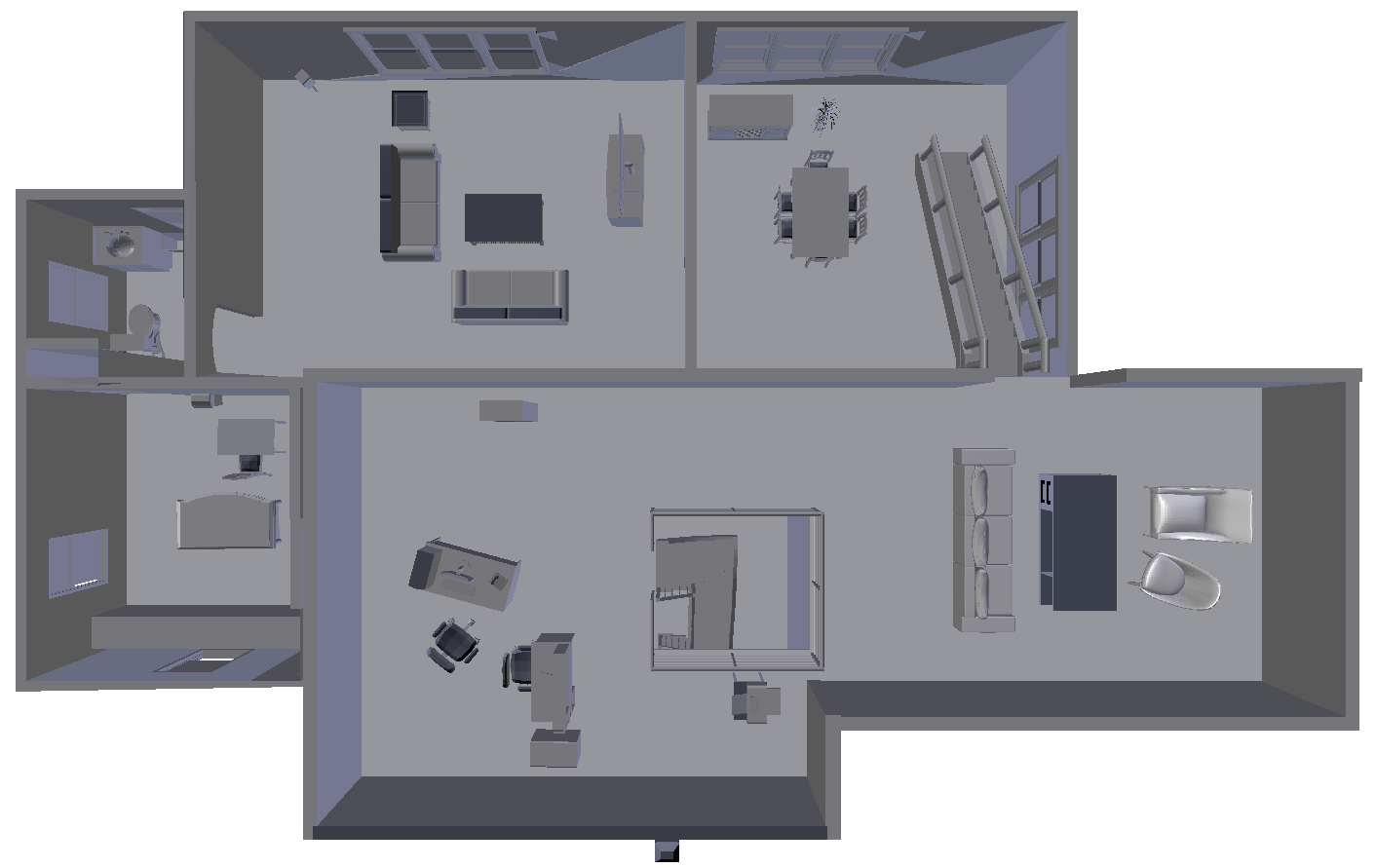}
        \label{fig:top_floor}
    }
    \caption{Top-down view of only the bottom floor~\protect\subref{fig:bottom_floor} and the top floor (bottom right) with the bottom floor~\protect\subref{fig:top_floor} of the testing environment. The environment is approximately $23.3\text{m} \times 15.12\text{m} \times 7.13\text{m}$.}
    \label{fig:environment}
\end{figure}

\section{Experiments and Results}
\label{sec:experiments}

To test the capabilities of the footstep planner\footnote{Our implementation is available at \url{https://github.com/vinitha910/homotopy_guided_footstep_planner}.} with various heuristics, we task a Poppy humanoid robot~\cite{poppy} to plan footstep motions to a goal region in a house environment.
We run our planner on 2 types of queries varying in their degree of complexity and evaluate the performance of our footstep planner by comparing the overall planning times (heuristic computation times and planning times) when using 3 different sets of heuristic functions:
\begin{enumerate}
	\item []\begin{set} One backward 2D Dijkstra heuristic $\mathcal{H}_{\text{Dijk}}$. \label{hset:s1}\end{set}
	\item []\begin{set} One backward 2D Dijkstra heuristic $\mathcal{H}_{\text{Dijk}}$
			and
		  one homotopy-based heuristic~$\mathcal{H}_s$.\label{hset:s2}\end{set}
	\item []\begin{set} One backward 2D Dijkstra heuristic $\mathcal{H}_{\text{Dijk}}$
			and
		  three homotopy-based heuristics $\{ \mathcal{H}_s \vert s \in \{s_1, s_2, s_3 \} \}$.\label{hset:s3}\end{set}
\end{enumerate}

\subsection{Test Setup}

\begin{figure}[!t]
    \centering
    \includegraphics[width=0.7\textwidth]{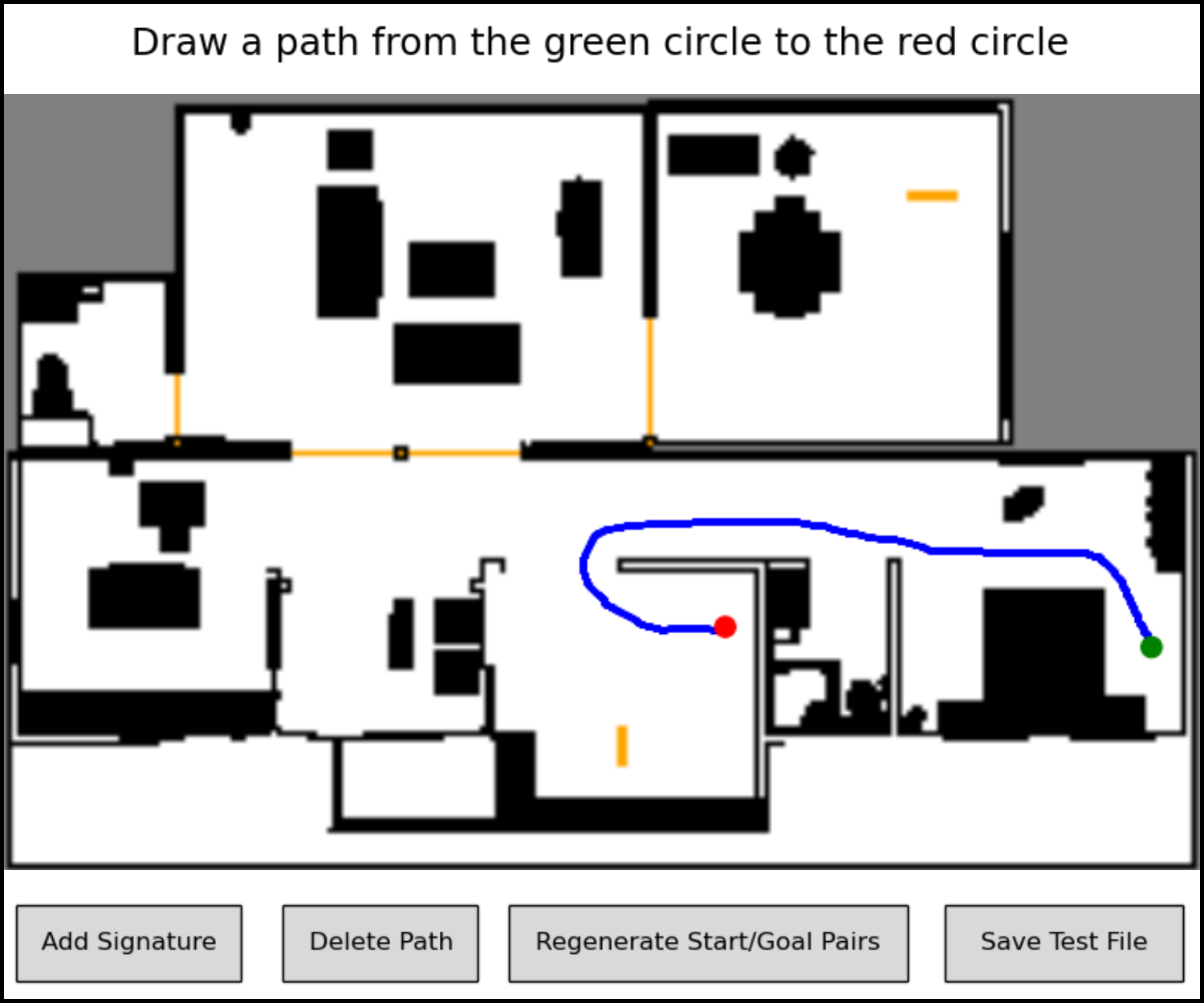}
    \caption{Example of interface to provide reference paths for constructing homotopy-based heuristics.}
    \label{fig:gui}
\end{figure}

We run our footstep planner in a two-story house environment(\figref{fig:environment}) with each set of heuristic functions on 40 \textit{simple} and 40 \textit{complex} queries. Each of the queries have randomly generated start and goal configurations. The path between the configurations in the simple queries are not obstructed by narrow passages. The path between the configurations in the complex queries, on the other hand, are obstructed by at least one narrow passage that the robot may or may not be able to pass through. Additionally, our planner used a discretization of $0.1$m, $w_1 = 3.0$ (heuristic inflation value)~\cite{aine14mha,ASNHL16} and $w_2 = 2.0$ (inadmissible search prioritization factor)~\cite{aine14mha,ASNHL16}.

To generate our homotopy-based heuristic functions we developed an interface that displays a projection of the inflated obstacles in the environment as well as the start (red circle) and goal configuration (green circle) (\figref{fig:gui}). The user can then draw paths from the goal to start configuration. All the signatures for the the reference paths are automatically generated and are used to compute the homotopy-based heuristic functions.

\subsection{Simple Queries}

\begin{figure}[t!]
    \centering
    \subfloat[Heuristic Computation Times]{
        \includegraphics[width=0.5\columnwidth]{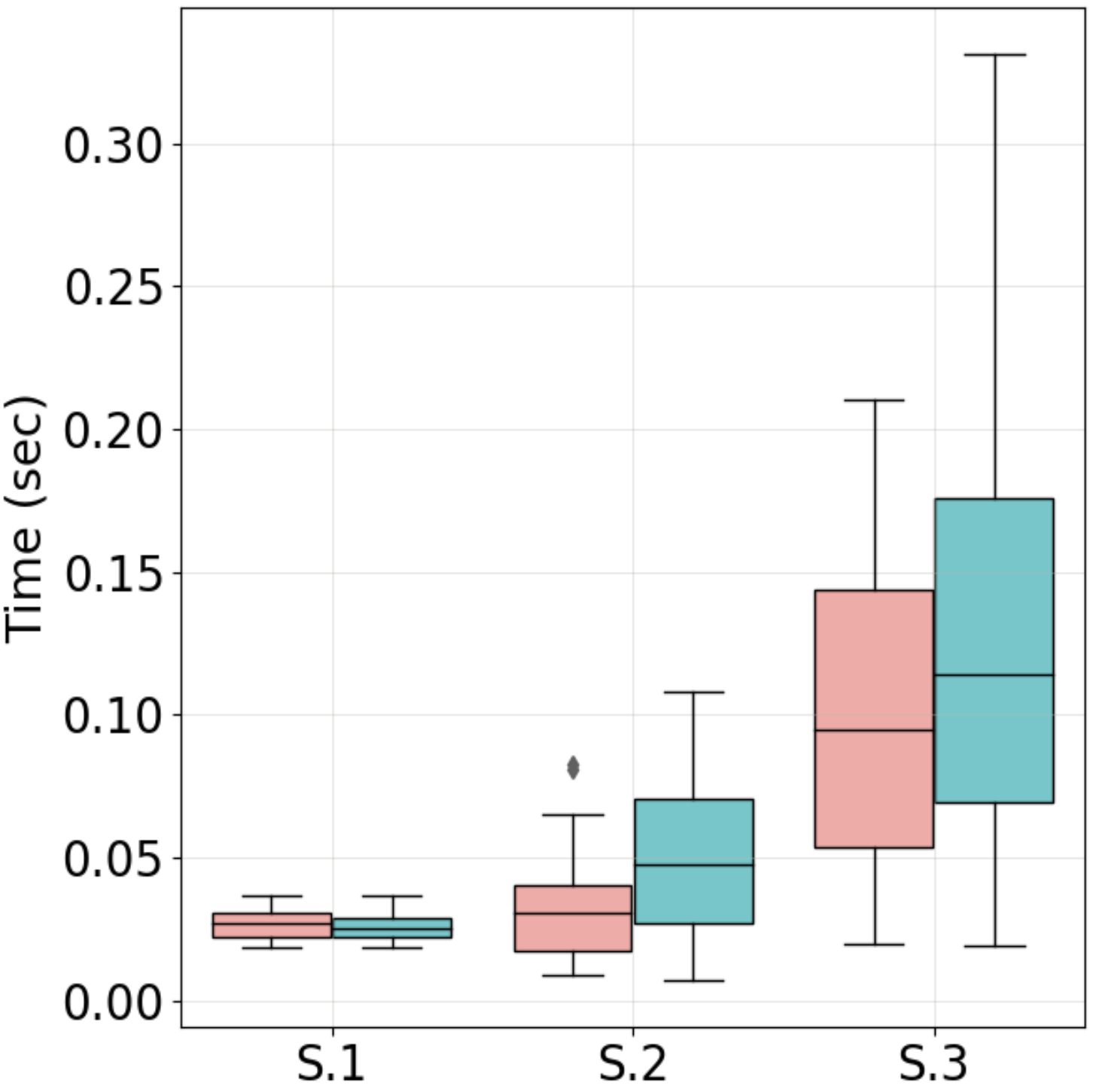}
        \label{fig:heuristic_computation_times}
    }
    \subfloat[Overall Planning Times]{
        \includegraphics[width=0.5\columnwidth]{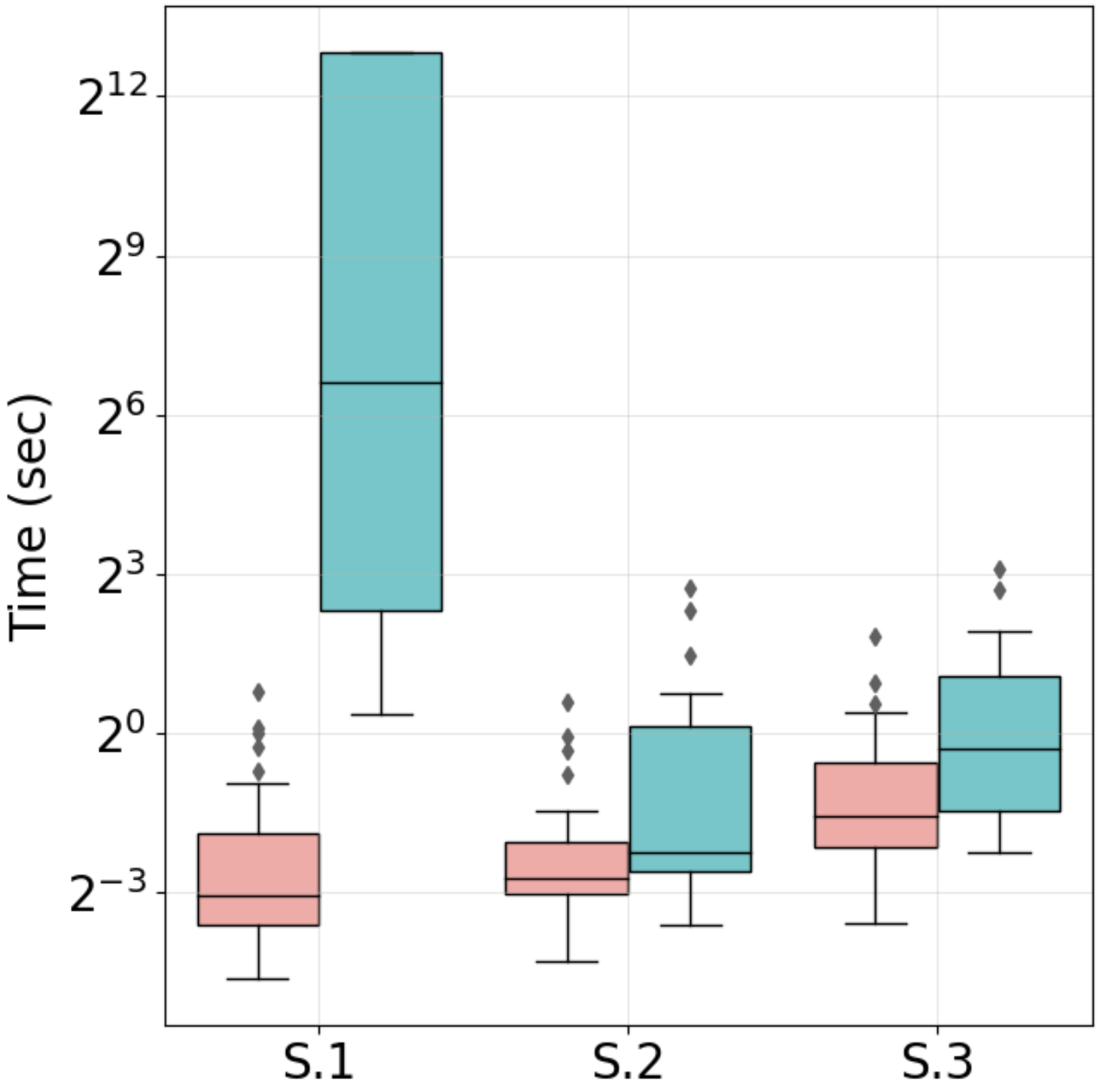}
        \label{fig:planning_times}
    }
    \caption{Box plots of heuristic computation time~\protect\subref{fig:heuristic_computation_times} and planning time~\protect\subref{fig:planning_times} for the different sets of  heuristic functions \hset{hset:s1}-\hset{hset:s3} for easy~(\protect\CaptionCircle{trivialbox}) and complex~(\protect\CaptionCircle{complexbox}) queries. The middle line in the box plot represents the median.
    We can see~\protect\subref{fig:heuristic_computation_times} that computing heuristics for \hset{hset:s2} and \hset{hset:s3} incurs slightly more computation time when compared to \hset{hset:s1}.
    However, this only slightly increases the planning time while for complex queries yields a speedup by several orders of magnitude (see also Fig.~\ref{fig:speedup}).
    Notice that $y$-axis (time) is in logarithmic scale for~\protect\subref{fig:planning_times}.}
    \label{fig:boxplots}
\end{figure}

In the simple queries the overall performance of the footstep planner for all sets of heuristic functions is comparable. The time taken to compute the homotopy-based heuristic functions is slightly longer than that of~$\mathcal{H}_{\text{Dijk}}$. Therefore, the overall planning time while using \hset{hset:s2} or \hset{hset:s3} is either similar to or slightly more than the planning time when using \hset{hset:s1} (See~\figref{fig:boxplots}).

\subsection{Complex Queries}

In the complex queries there was a speedup in planning of several orders of magnitude when using \hset{hset:s2} or \hset{hset:s3} in comparison to \hset{hset:s1}. While it took longer to compute \hset{hset:s2} and \hset{hset:s3}, these times were effectively negligible when comparing the planning times to when \hset{hset:s1} was used (See~\figref{fig:boxplots}).

\begin{figure}[t!]
    \centering
    \subfloat[]{
        \includegraphics[width=0.5\columnwidth]{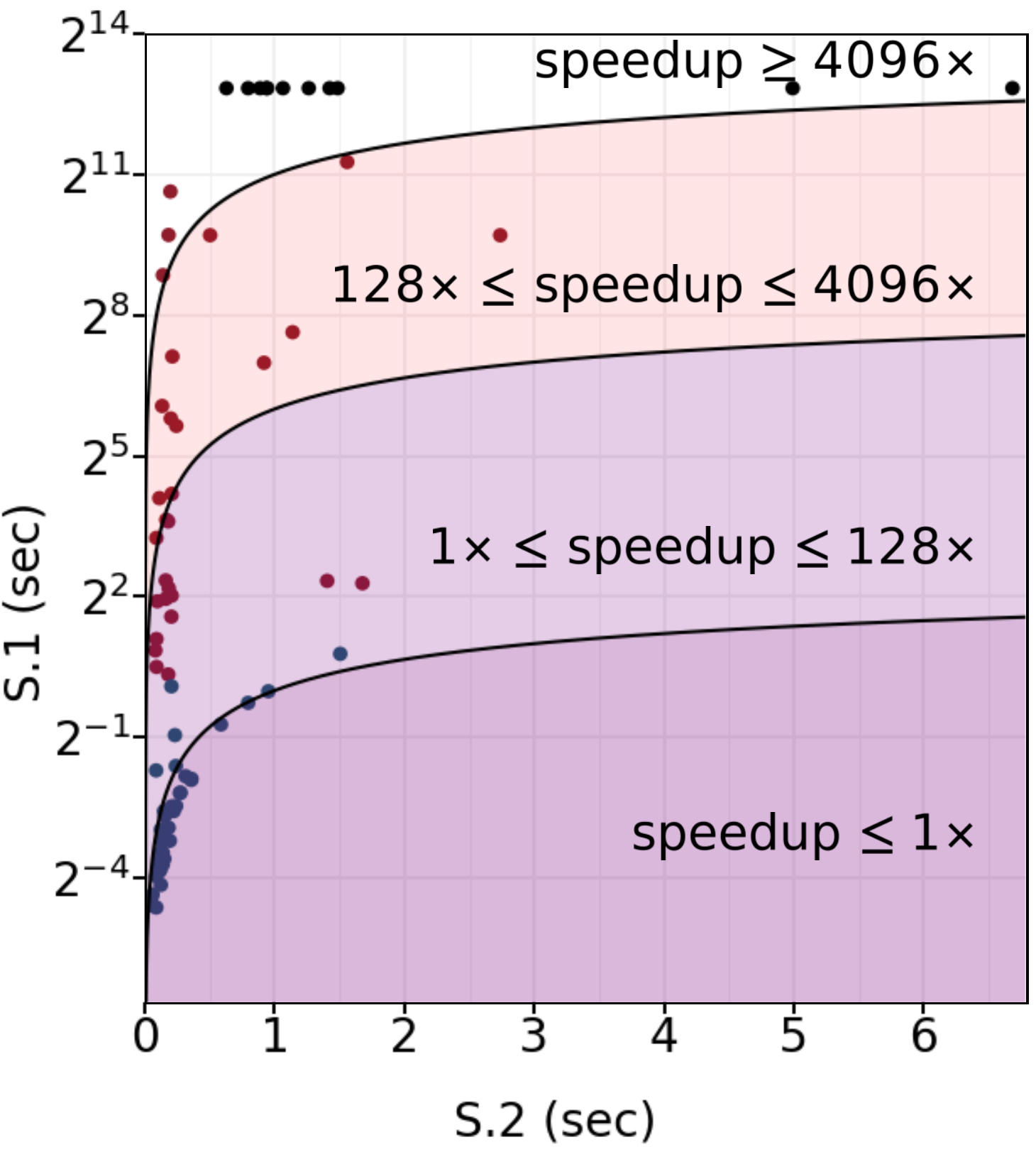}
        \label{fig:dijkstra_vs_hbsp_1}
    }
    \subfloat[]{
        \includegraphics[width=0.5\columnwidth]{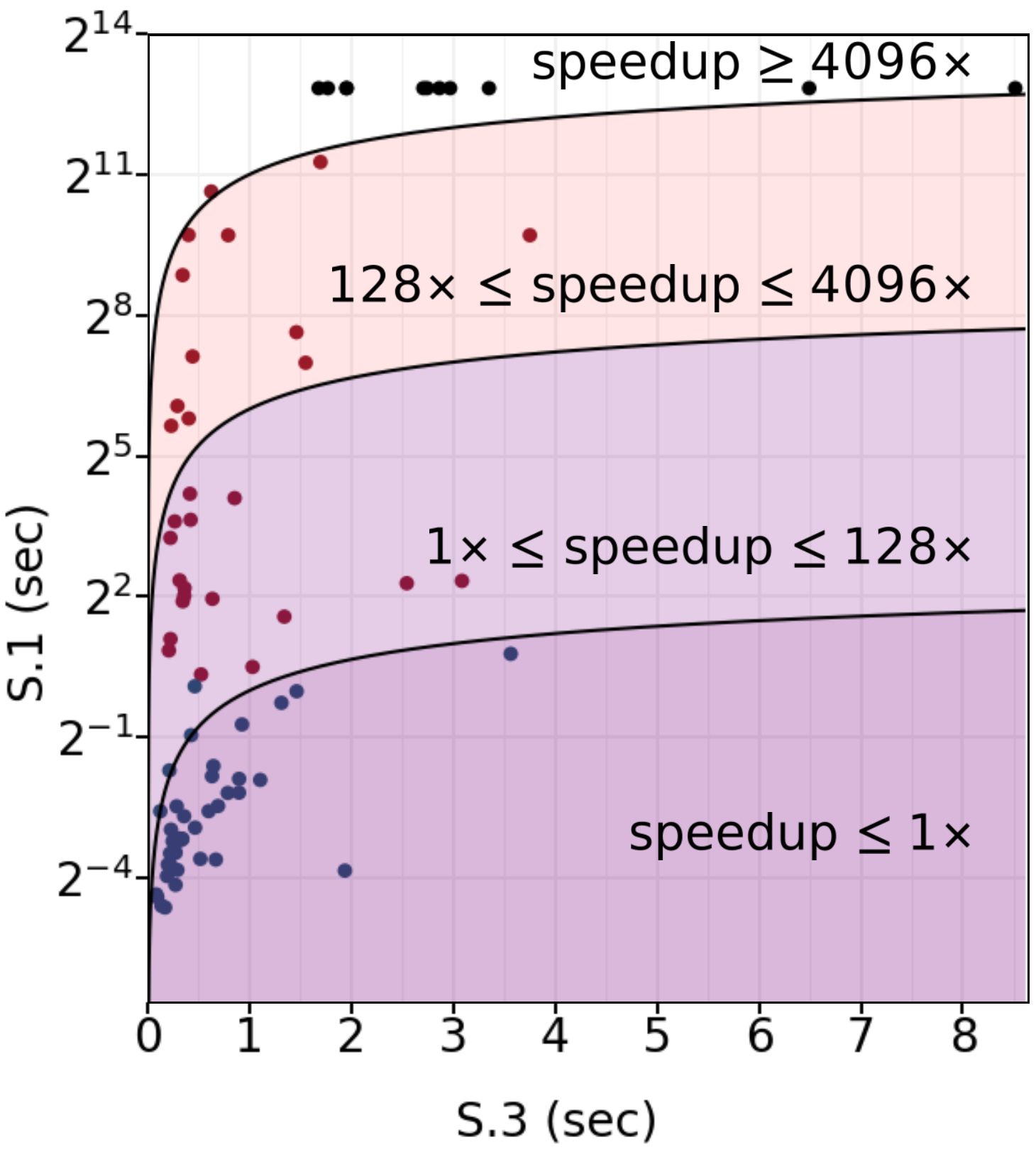}
        \label{fig:dijkstra_vs_hbsp_3}
    }
    \caption{Speedup in overall planning time (search time and heuristic computation) between \hset{hset:s2} and \hset{hset:s1}~\protect\subref{fig:dijkstra_vs_hbsp_1} and \hset{hset:s3} and \hset{hset:s1}~\protect\subref{fig:dijkstra_vs_hbsp_3}.
    The running times for simple~(\protect\CaptionCircle{trivialscatter}) queries is comparable among all algorithms (all points are close to the line speedup $\leq 1 \times$).
    However, in complex~(\protect\CaptionCircle{complexscatter}) queries there is a speed up in planning times of \textbf{16 to more than 4096 times} when using \hset{hset:s2} or \hset{hset:s3}. There are a few complex~(\protect\CaptionCircle{verycomplexscatter}) queries where the planner cannot find a solution with the allotted memory (16 GB) when using \hset{hset:s1} but can quickly find a solution when using \hset{hset:s2} or \hset{hset:s3}. Notice that $y$-axis is in logarithmic scale.}
    \label{fig:speedup}
\end{figure}

Additionally, when using \hset{hset:s2} there were several scenarios where there was a \textbf{16 to 2048 times} speedup in planning times and a few scenarios with more than \textbf{2048 times} speedup in planning times~(\subfigref{fig:speedup}{fig:dijkstra_vs_hbsp_1}). When using \hset{hset:s3} there are scenarios with less speedup in planning times in comparison to \hset{hset:s2}~(\subfigref{fig:speedup}{fig:dijkstra_vs_hbsp_3}) due to the poor quality of some of the heuristics. However, with at least one informative heuristic function for each scenario, the footstep planner was able to very quickly guide the search to the goal. Furthermore, there were some scenarios where, when using \hset{hset:s1}, the planner could not find a solution with the allotted memory (16 GB), however both \hset{hset:s2} and \hset{hset:s3} were able to quickly find a solution.

\section{Discussion}
\label{sec:future}

We presented an approach for automatically generating heuristic functions given user-defined homotopy classes in 2D and 2.5D environments to effectively plan footstep motions for a humanoid robot. We showed that generating informative heuristics can significantly reduce planning times. Additionally, we presented an approach that allows users to easily construct these heuristics. 

Our experiments showed that in \textit{simple} queries the performance of the footstep planner, when using the heuristics generated through our approach, is comparable to that of the baseline approach. However, in \textit{complex} queries we showed that when the footstep planner uses the heuristics generated through our approach, it plans several orders of magnitude faster than the baseline approach. Additionally, we showed that providing the footstep planner heuristic functions of poor quality does impede its performance. 


We believe there are three promising directions that may improve the quality of our footstep planner. First, we can automatically generate homotopy classes for our heuristic functions to allow for a fully autonomous planner. Second, we can use our approach only when the planner ceases to make progress towards the goal~\cite{ISL17}. For example, in \textit{simple} queries our approach does not improve planning times; it only has an impact on \textit{complex} queries. Therefore we can utilize our approach only when the planner gets stuck to minimize planning and heuristic computation times. Finally, our approach requires that the environment be projected onto a series of surfaces. Currently, these surfaces are defined defined by the user but ideally we would like them to be automatically generated.  

\section*{Acknowledgments}
\label{sec:acknowledgments}

This work was supported by National Science Foundation Grants IIS-1659774, IIS-1409549 and DGE-1762114.






\bibliography{references}
\bibliographystyle{elsarticle-num}

\end{document}